\documentclass{article} % For LaTeX2e
\usepackage{iclr2024_conference,times}

\usepackage{amsmath,amsfonts,bm}

\def\eqref#1{equation~\ref{#1}}
\def\1{\bm{1}}

\DeclareMathAlphabet{\mathsfit}{\encodingdefault}{\sfdefault}{m}{sl}
\SetMathAlphabet{\mathsfit}{bold}{\encodingdefault}{\sfdefault}{bx}{n}

\usepackage{hyperref}
\usepackage{url}
\usepackage{paralist}
\usepackage{wrapfig}

\usepackage{booktabs}
\usepackage{tablefootnote}
\usepackage{amsmath}
\usepackage{enumitem}
\usepackage{algpseudocode}
\usepackage{xspace}
\usepackage{subcaption}
\usepackage{multirow}
\usepackage{multicol}
\usepackage{xcolor}
\usepackage{amsthm}
\usepackage{amssymb}

\usepackage{bbm}
\usepackage{tabularx, booktabs}
\usepackage{graphicx}
\newcolumntype{C}[1]{>{\centering\let\newline\\\arraybackslash\hspace{0pt}}m{#1}}
\newcolumntype{R}[1]{>{\raggedleft\let\newline\\\arraybackslash\hspace{0pt}}m{#1}}
\newcolumntype{L}[1]{>{\raggedright\let\newline\\\arraybackslash\hspace{0pt}}m{#1}}
\usepackage{colortbl}

\usepackage[linesnumbered,ruled,vlined]{algorithm2e}
\SetAlgoNlRelativeSize{-1}

\newcommand*\samethanks[1][\value{footnote}]{\footnotemark[#1]}

\usepackage{pifont}
\definecolor{darkgreen}{rgb}{0.0, 0.5, 0.0}
\newcommand*\colourcheck[1]{%
  \expandafter\newcommand\csname #1check\endcsname{\textcolor{#1}{\ding{52}}}%
}
\colourcheck{darkgreen}

\hypersetup{
     colorlinks   = true,
     citecolor    = blue
}
\newcommand{\purple}[1]{\textcolor{purple}{#1}}
\newcommand{\red}[1]{\textcolor{red}{#1}}
\newcommand{\blue}[1]{\textcolor{blue}{#1}}
\newcommand{\green}[1]{\textcolor{olive}{#1}}

\newcommand{\james}[1]{\red{[James:] #1}}
\newcommand{\christos}[1]{\green{[Christos:] #1}}
\newcommand{\jeremy}[1]{\blue{[Jeremy:] #1}}
\newcommand{\xiangsx}[1]{\purple{[Xiang:] #1}}

\renewcommand{\james}[1]{}
\renewcommand{\christos}[1]{}
\renewcommand{\jeremy}[1]{}
\renewcommand{\xiangsx}[1]{}

\newcommand{\bit}{\begin{compactitem}}
\newcommand{\eit}{\end{compactitem}}
\newcommand{\ben}{\begin{compactenum}}
\newcommand{\een}{\end{compactenum}}

\newtheorem{definition}{Definition}
\newtheorem{lemma}{Lemma}

\newcommand{\cat}[0]{\mathbin\Vert}

\newcommand{\doconcat}[1]{%
  \vcenter{#1\kern.2ex\hbox{$\Vert$}\kern.2ex}}

\newcommand{\hide}[1]{}

\newcommand{\method}{\textsc{NetInfoF}\xspace}
\newcommand{\prefix}{\textsc{NetInfoF}}
\newcommand{\analysis}{\textsc{\prefix\textunderscore Probe}\xspace}
\newcommand{\model}{\textsc{\prefix\textunderscore Act}\xspace}
\newcommand{\score}{\textsc{\prefix\textunderscore Score}\xspace}

\newcommand{\myEmph}[1]{{\em #1}}

\newcommand{\general}{ \myEmph{General}\xspace}
\newcommand{\principled}{ \myEmph{Principled}\xspace}
\newcommand{\effective}{ \myEmph{Effective}\xspace}
\newcommand{\scalable}{ \myEmph{Scalable}\xspace}

\newcommand{\robust}{ \myEmph{Robust}\xspace}

\newcommand{\emphasize}[1]{\textbf{\underline{\smash{#1}}}}

\newtheorem{theorem}{Theorem}

\newcommand{\pOne}{{p_1}}
\newcommand{\pTwo}{{p_2}}
\newcommand{\pEye}{{p_i}}
\newcommand{\pMax}{{p_{max}}}
\newcommand{\pEn}{{p_n}}

\title{\method Framework: Measuring and\\Exploiting Network Usable Information}

\author{
Meng-Chieh Lee\textsuperscript{\textnormal{1,}}\thanks{The work is done while being an intern at Amazon.} , 
Haiyang Yu\textsuperscript{\textnormal{2}}, 
Jian Zhang\textsuperscript{\textnormal{3}}, 
Vassilis N. Ioannidis\textsuperscript{\textnormal{3}},
Xiang Song\textsuperscript{\textnormal{3}}, \\
\textbf{
Soji Adeshina\textsuperscript{\textnormal{3}}, 
Da Zheng\textsuperscript{\textnormal{3,}}\thanks{Corresponding authors.} , 
Christos Faloutsos\textsuperscript{\textnormal{1,3,}}\samethanks
} \\
\textsuperscript{1}Carnegie Mellon University, 
\textsuperscript{2}Texas A\&M University,
\textsuperscript{3}Amazon \\
\texttt{\{mengchil,christos\}@cs.cmu.edu}, 
\texttt{\{haiyang\}@tamu.edu}, \\
\texttt{\{jamezhan,ivasilei,xiangsx,adesojia,dzzhen\}@amazon.com} \\
}

\iclrfinalcopy % Uncomment for camera-ready version, but NOT for submission.
\begin{document}

\maketitle

\begin{abstract}
Given a node-attributed graph, and a graph task (link prediction or node classification),
can we tell if a graph neural network (GNN) will perform well?
% Graph tasks, such as link prediction and node classification, are commonly solved by adopting GNNs.
More specifically, do the graph structure and the node features carry enough usable information for the task?
% Sometimes, the node features may be useless to forecast links; some other times, the graph structure may be useless.
Our goals are 
(1) to develop a fast tool to measure how much information is in the graph structure and in the node features, and
(2) to exploit the information to solve the task, if there is enough.
% However, they may not always work as expected in some graph scenarios, for example, in the case that the structure is randomly connected. 
% Figuring out the reasons is not trivial and can cost a huge amount of time and resources. 
% More formally, we want to quantify the usable information in the graph data. 
We propose \method, a framework including \analysis and \model, for the measurement and the exploitation of network usable information (NUI), respectively.
Given a graph data, \analysis measures NUI without any model training, and \model solves link prediction and node classification, while two modules share the same backbone.
In summary, \method has following notable advantages:
(a) \general, handling both link prediction and node classification;
(b) \principled, with theoretical guarantee and closed-form solution;
(c) \effective, thanks to the proposed adjustment to node similarity;
(d) \scalable, scaling linearly with the input size.
In our carefully designed synthetic datasets, \method correctly identifies the ground truth of NUI and is the only method being robust to all graph scenarios.
Applied on real-world datasets, \method wins in \textit{11 out of 12} times on link prediction compared to general GNN baselines.

% Given a node-attributed graph, and a graph task (link prediction or node classification), can we tell if a graph neural network (GNN) will perform well? More specifically, do the graph structure and the node features carry enough usable information for the task? Our goals are (1) to develop a fast tool to measure how much information is in the graph structure and in the node features, and (2) to exploit the information to solve the task, if there is enough. We propose NetInfoF, a framework including NetInfoF_Probe and NetInfoF_Act, for the measurement and the exploitation of network usable information (NUI), respectively. Given a graph data, NetInfoF_Probe measures NUI without any model training, and NetInfoF_Act solves link prediction and node classification, while two modules share the same backbone. In summary, NetInfoF has following notable advantages: (a) General, handling both link prediction and node classification; (b) Principled, with theoretical guarantee and closed-form solution; (c) Effective, thanks to the proposed adjustment to node similarity; (d) Scalable, scaling linearly with the input size. In our carefully designed synthetic datasets, NetInfoF correctly identifies the ground truth of NUI and is the only method being robust to all graph scenarios. Applied on real-world datasets, NetInfoF wins in 11 out of 12 times on link prediction compared to general GNN baselines.
\end{abstract}

\section{Introduction}
Given a graph with node features, how to tell if a graph neural network (GNN) can perform well on graph tasks or not?
How can we know what information (if any) is usable to the tasks, namely, link prediction and node classification?
% We also want to know when we should apply.
GNNs \citep{kipf2017semisupervised,hamilton2017inductive,velickovic2018graph} are commonly adopted on graph data to generate low-dimensional representations that are versatile for performing different graph tasks.
% , such as link prediction and node classification. 
\hide{
 However, due to the complicated nature of node-attributed graphs, GNNs may not work as expected in some scenarios. 
 For example, if the structure of the given graph is uncorrelated with the task labels, then methods like logistic regression (LogitReg) or multi-layer perceptron (MLP) are better solutions than GNNs \citep{liu2021non,Lim21LINKX}.
 % \james{do we have references(citations) about this phenomena?} 
 Moreover, in real-world scenarios, the graphs are extremely large, making the time and resource consumption prohibitive for analyzing GNNs by training them with different settings.
% \james{may say time and resource consuming}
}% end hide
However, sometimes there are no network effects,
and training a GNN will be a waste of computation time.
% It is thus essential to determine the necessity of training GNNs with the aim of resource conservation. 
That is to say, we want a measurement of how informative the graph structure and node features are for the task at hand, which we call \textit{network usable information (NUI)}.

% Informally, we want an estimate of how informative is the
% connectivity (structure) and for the task at hand (link prediction, etc). We call this concept {\em usable network information}
% and we give a formal measurement later (see Eq(2)***)

%To achieve this goal, a preliminary analysis tool to measure the usable information in the given graph data, which we call \textit{network usable information (NUI)}, is desired.
%This is exactly one of our contributions: We propose to measure it using **netInfo-score* as in Eq. ***
\hide{
There are several categories of methods to measure the usable information. 
One typical way to measure the usable information of a given data is by estimating its mutual information \citep{kraskov2004estimating} with the targets.
% , which is used for sequential feature selection \citep{li2017feature}.
Another branch of studies \citep{Xu2020A, ethayarajh2022understanding} turns to analyze the usable information with trained models.
% Nevertheless, both of them suffer from some serious obstacles in practice: 
Nevertheless, the former suffers from the difficulty for the exact computation, and the latter is impotent without model training.
% \haiyang{The obstacles exist in both branches or each one accordingly?}
Furthermore, they cannot be used directly to measure the usable information in the graph data.
The graph data exists in a non-Euclidean space, consisting of more than one component, such as node features and graph structure.
% \xiangsx{we should be more specific, Let's say Unlike other kinds of data such as tabular data or image data}
% Unlike other kinds of data (e.g. tabular or image data), the graph data has information from at least two different components, i.e., graph structure and node features.
Although GNNs can be used to encode the graph data, there are some drawbacks. 
% \haiyang{I am not sure about the correctness of this sentence. Feel free to reorganize it with specific details using the GNNs for graph data to measure usable information.} 
While the node embeddings of GNNs are shown to be informative by propagating the node features through the graph structure, they are not available without model training.
Additionally, the information of each individual component remains unclear.
% Thus, it is important to develop scalable methods to measure the usable information for graph data considering each component.
}% end hide

% \textcolor{blue}{There is no definition of NUI in either Introduction or Problem definition (section 3.1). People may wonder what does NUI exactly mean.}

\begin{figure}[t]
\begin{minipage}[b] {0.4\linewidth}
    \centering
    \includegraphics[height=1.65in]{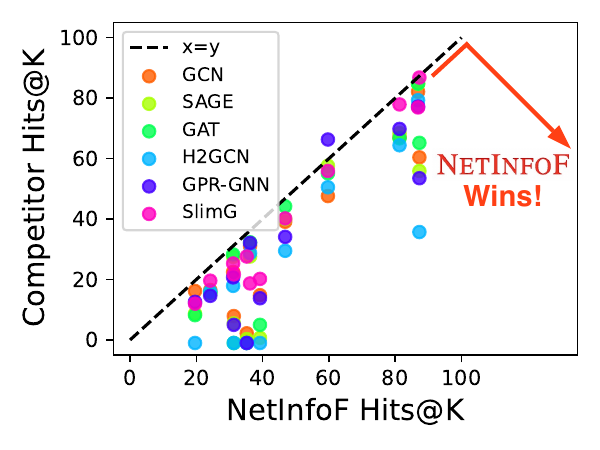}
    \vspace{-9mm}
    \caption{\emphasize{\method wins} in real-world datasets on link prediction (most points are below or on line $x=y$). \label{fig:cj1}}
\end{minipage} \hfill
\begin{minipage}[b] {0.58\linewidth}
    \centering
    \subfloat[Link Prediction]{\vspace{-3mm}\includegraphics[height=1.45in]{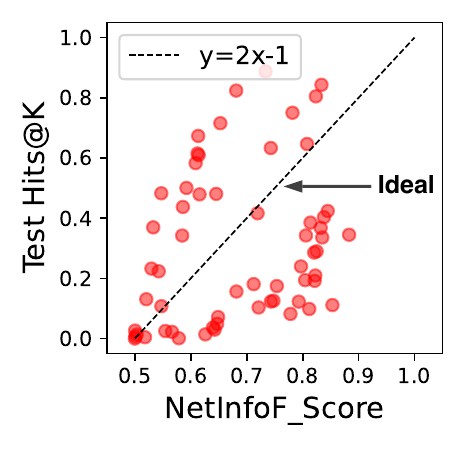}}
    \subfloat[Node Classification]{\vspace{-3mm}\includegraphics[height=1.45in]{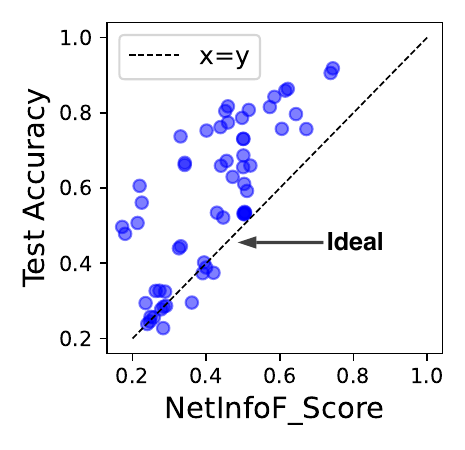}} 
    \vspace{-3mm}
    \caption{\emphasize{\score highly correlates to test}\xspace\emphasize{performance} in real-world datasets.
    Each point denotes the result of a component from each dataset. 
    \label{fig:accnis_real}}
\end{minipage}
\vspace{-5mm}
\end{figure}

We propose \method, a framework to measure and exploit NUI in a given graph.
% \textit {network usable information (NUI)}.
First, \analysis measures NUI of the given graph with \score (Eq.~\ref{eqn:score}), which we proved is lower-bound the accuracy (Thm.~\ref{thm:t2}).
Next, our \model solves both the link prediction and node classification by sharing the same backbone with \analysis.
To save training effort, we propose to compute \score by representing different components of the graph with carefully derived node embeddings.
For link prediction, we propose the adjustment to node similarity with a closed-form formula to address the limitations when the embeddings are static. 
% i.e., they can not (1) precisely measure the node similarity when the neighbors have heterophily embeddings, and (2) well distinguish between the negative and positive edges.
We demonstrate that our derived embeddings contain enough usable information, by showing the superior performance on both tasks.
In Fig.~\ref{fig:cj1}, \model outperforms the GNN baselines most times on link prediction;
in Fig.~\ref{fig:accnis_real}, \score measured by \analysis highly correlates to the test performance in real-world datasets. 
% \haiyang{In Fig.~\ref{fig:accnis_real} (a), the relationship between Test Hits@100 and \score seems to be not that correlated. Some are above the RG while some are below it. I guess the expected results are all the results are above the RG. If so, this figure may need to be justified.}

In summary, our proposed \method has following advantages:
\vspace{-1mm}
\ben
    \item \textbf{\general}, handling both node classification and link prediction (Lemma~\ref{lem:cheapcm}-\ref{lem:expcm});
    \item \textbf{\principled}, with theoretical guarantee (Thm.~\ref{thm:t1}-\ref{thm:t2}) and closed-form solution (Lemma~\ref{lem:cheapcm}-\ref{lem:expcm}); 
    \item \textbf{\effective}, thanks to the proposed adjustment of node similarity (Fig.~\ref{fig:cj1});
    % winning in $11$ out of $12$ real-world datasets on link prediction (Fig.~\ref{fig:cj1}); 
    \item \textbf{\scalable}, scaling linearly with the input size (Fig.~\ref{fig:scalable}).
\een
\vspace{-1mm}
In synthetic datasets, \method correctly identifies the ground truth of NUI and is the only method being robust to all possible graph scenarios;
in real-world datasets, \method wins in \textit{11 out of 12} times on link prediction compared to general GNN baselines.

\vspace{-1mm}
{\bf Reproducibility:} Code is at \url{https://github.com/amazon-science/Network-Usable-Info-Framework}.

% \begin{figure}[t]
% \begin{minipage}[b] {0.275\linewidth}
%     \centering
%     \includegraphics[height=1.38in]{FIG/comparison.pdf} 
%     \caption{\emphasize{\method wins} in real-world datasets on link prediction (most points are below line $x=y$). \label{fig:cj1}}
% \end{minipage} \hfill
% \begin{minipage}[b] {0.71\linewidth}
%     \centering
%     \subfloat[Link Prediction]{\includegraphics[height=1.2in]{FIG/ni_lp_hit_real.pdf}
%     \includegraphics[height=1.2in]{FIG/ni_lp_roc_real.pdf}}
%     \rulesep
%     \subfloat[Node Classification]{\includegraphics[height=1.2in]{FIG/ni_nc_real.pdf}} 
%     \caption{\emphasize{\score correlates to test performance}. Each point denotes a component from real-world datasets averaged from five splits. RG stands for random guess, depicted with gray cross hatching.\label{fig:accnis_real}}
% \end{minipage}
% \end{figure}

\vspace{-3mm}
\section{Related Works}
\vspace{-3mm}
\begin{wrapfigure}{L}{0.52\textwidth}
\vspace{-5mm}
\begin{minipage}[h] {0.99\linewidth}
\captionof{table}{\emphasize{\method matches all properties}, while baselines miss more than one property. \label{tab:salesman}}
\vspace{-3mm}
\centering{\resizebox{1\columnwidth}{!}{
    \begin{tabular}{ l | l | ccc | c }
        \hline
        \multicolumn{2}{c|}{\bf Property} & 
        \rotatebox{90}{General GNNs} &
        \rotatebox{90}{Subgraph GNNs} &
        \rotatebox{90}{SlimG} & 
        \rotatebox{90}{\bf \method} \\ 
        \hline
        {\bf 1. \general} & 
        1.1. Node Classification & \darkgreencheck &  & \darkgreencheck & \darkgreencheck \\ 
        {~~~~~w.r.t. Graph Task} & 
        1.2. Link Prediction & \darkgreencheck & \darkgreencheck &  & \darkgreencheck \\ 
        \hline
        \multirow{2}{*}{\bf 2. \principled} & 
        2.1. Theoretical Guarantee &  &  &  & \darkgreencheck \\
         & 
        2.2. Closed-Form Solution &  &  & \darkgreencheck & \darkgreencheck \\
        \hline
        \multicolumn{2}{l|}{\bf 3. \scalable} & \darkgreencheck &  & \darkgreencheck & \darkgreencheck \\
        \hline
        % \multicolumn{2}{l|}{\bf 4. \robust} &  &  & \darkgreencheck & \darkgreencheck \\
        {\bf 4. \robust} & 
        4.1. Node Classification &  &  & \darkgreencheck & \darkgreencheck \\ 
        {~~~~~w.r.t. Input Scenario} & 
        4.2. Link Prediction &  &  &  & \darkgreencheck \\ 
        \hline
    \end{tabular}
}}
\end{minipage}
\vspace{-5mm}
\end{wrapfigure}

We introduce the related work in two groups: information theory, and GNNs.
%; and then introduce graph neural networks (GNNs), along with its subcategory linear GNNs.
In a nutshell, \method is the only one fulfills all the properties as shown in Table~\ref{tab:salesman}.

\textbf{Information Theory.}
% \paragraph{Information Theory.}
The typical measure of the dependence between the random
variables is the mutual information \citep{kraskov2004estimating}.
It is powerful and widely used in sequential feature selection \citep{li2017feature}, but its exact computation is difficult
% suffers from many well-known difficulties for the exact computation 
\citep{paninski2003estimation, belghazi2018mutual}
especially on continuous random variables \citep{ross2014mutual, mesner2020conditional} 
% and lack of universal stopping rules 
and high-dimensional data
\citep{franccois2006permutation, mielniczuk2019stopping}.
% which exacerbates when the number of dimensions is large.
% Another branch of studies
Recently \citep{Xu2020A, ethayarajh2022understanding} proposed the concept of $\mathcal{V}$-information.
% by analyzing the information in the output of a trained model.
However, the definition needs a trained model,
which is expensive to obtain and is dependent on the quality of training.
\hide{
 This bypasses the traditional difficulties, and works fine when the raw data (e.g. text) is hard to connect with the targets without feature extraction.
 However, training the model raises an extra computation overhead, as well as a huge uncertainty to the analysis because of model tuning.
 In this case, the data can be seen as useless if a model is not properly tuned, especially when the model is complex \citep{alain2017understanding}.
 Moreover, when the trained model is available, some metrics, like accuracy widely used in linear probing \citep{kiros2015skip, chen2020simple}, can be more intuitive than $\mathcal{V}$-information in most cases. 
}% end hide

% Jeremy: maybe this can be moved to the method section
% Only a few works study the usable information in the graphs, and they are not feasible in our problem settings because of three challenges.
% Moreover, this concept suffers from the following issues, for our setting:
Only a few works study the usable information in the graphs, but are not feasible in our problem settings because of three challenges, i.e., our desired method has to:
% (1) the usable information can not be measured without training any models (see \citep{akhondzadeh2023probing}).
(1) work without training any models, where \citet{akhondzadeh2023probing} requires model training;
% \citet{akhondzadeh2023probing} conducts linear probing to study the representational power of GNNs on graph properties; nevertheless, the model training and tuning is time-consuming.
(2) identify which components of the graph are usable, where \citet{pmlr-v202-dong23a} ignores the individual components; and
% proposes a metric to measure the structural noise level w.r.t. the node features, while the usefulness of individual components is ignored.
% For example, in the case that only the node features are usable to the task, one should just use MLP instead of GNNs.
% (2) it is unclear how to use it for link prediction
(3) generalize to different graph tasks, where
% the method should be able to generalize to different graph tasks, namely, link prediction and node classification. 
\citet{lee2022ultraprop} focuses on node classification only.

% \paragraph{Graph Neural Networks (GNNs).}
\textbf{Graph Neural Networks.}
% Various GNNs have shown their power in multiple scenarios. 
Although most GNNs learn node embeddings
% by message passing scheme \citep{gori2005new, scarselli2008graph}
assuming homophily,
% While in heterophily graphs, similar nodes may be structurally distant from each other.
some GNNs \citep{abu2019mixhop, zhu2020beyond, chien2021adaptive, liu2021non} break this assumption 
by handling $k$-step-away neighbors differently for every $k$, and some systematically study the heterophily graphs on node classification \citep{platonov2022characterizing, luan2022revisiting, luan2023graph, mao2023demystifying, chen2023exploiting, ma2021homophily}.
% consider the distant nodes when aggregating information.
% by treating neighbors from different steps differently.
% While most works are evaluated on node classification, some other works focus on link prediction.
% \citet{wang2022flashlight} uses GNNs as the encoder and studies the performance of different link decoders.
% GNNs can serve as a powerful tools on multiple tasks including node classification and link prediction.
% While GNNs can be generally used to solve multiple tasks including node classification and link prediction, 
Subgraph GNNs \citep{zhang2018link, yin2022algorithm} are designed only for link prediction and are expensive in inference.
% They extract the subgraph to better infer the links, which results in a tremendous training and inference overhead, and are not considered in this paper.
On the other hand, linear GNNs \citep{wu2019simplifying, wang2021dissecting, zhu2021simple, li2022g, yoo2023less} target interpretable models. 
Such approaches remove the non-linear functions and maintain good performance.
% One of their advantages is that the node embeddings are derived instead of being learned, thus they are available before the model training.
% \citet{wu2019simplifying} reduces GCN \citep{kipf2017semisupervised} to closed-form matrix multiplication by removing its non-linear functions.
% \citet{wang2021dissecting, zhu2021si mple} manually assign the strength to the self-loop and the structure in order to propagate farther.
% \citet{li2022g} generalizes to heterophily graphs by concatenating derived features with different bandwidths.
% As the only method being robust to all scenarios, \textsc{SlimG} \citep{yoo2023less} achieves the outstanding performance by addressing all the pain points of GNNs.
As the only method being robust to all graph scenarios, \textsc{SlimG} \citep{yoo2023less} works well on node classification. However, it is unclear how well it works for link prediction.\looseness=-1
% \textcolor{red}{Cite and talk a little bit about SEAL.}

In conclusion, the proposed \method is the only one that fulfills all the properties in Table~\ref{tab:salesman}.
\label{sec:related}

\section{\score: Would a GNN work?}
How to tell whether a GNN will perform well on the given graph task?
A graph data is composed of more than one component, such as graph structure and node features.
In this section, we define our problem, and answer two important questions:
(1) How to measure the predictive information of each component in the graph?
(2) How to connect the graph information with the performance metric on the task?
We identify that a GNN is able to perform well on the task when its propagated representation is more informative than graph structure or node features.

% In this section, we first propose to represent the information of a graph by carefully deriving node embeddings that represent different components of the graph.
% We further derive theorems to connect the information and the performance.
% Based on the theorems and the derived embeddings, \method gives \score to each of the components, which is a novel metric for measuring NUI in graphs.
% \james{Can we also put these questions into the Introduction section. These questions are the key questions we want to address in this submission.}

\subsection{Problem Definition}
Given an undirected graph $G=(\mathcal{V}, \mathcal{E})$ with node features $\mathbf{X}_{|\mathcal{V}| \times f}$, where $f$ is the number of features, the problem is defined as follows:
\bit
\item \textbf{Measure} the network usable information (NUI), and
\item \textbf{Exploit} NUI, if there is enough, to solve the graph task.
\eit

We consider two most common graph tasks, namely link prediction and node classification.
In link prediction, $\mathcal{E}$ is split into $\mathcal{E}_{\text{train}}$ and $\mathcal{E}_{\text{pos}}$.
The negative edge set $\mathcal{E}_{\text{neg}}$ is randomly sampled with the same size of $\mathcal{E}_{\text{pos}}$.
The goal is to predict the existence of the edges, $1$ for the edges in $\mathcal{E}_{\text{pos}}$, and $0$ for the ones in $\mathcal{E}_{\text{neg}}$.
In node classification, $|\mathcal{V}_{\text{train}}|$ node labels $\mathbf{y} \in \{1, ..., c\}^{|\mathcal{V}_{\text{train}}|}$ are given, where $c$ is the number of classes.
The goal is to predict the rest $|\mathcal{V}| - |\mathcal{V}_{\text{train}}|$ unlabeled nodes' classes.

\subsection{Proposed Derived Node Embeddings}
To tell whether a GNN will perform well, we can analyze its node embeddings, but they are only available after training.
For this reason, we propose to analyze the derived node embeddings in linear GNNs.
More specifically, we derive $5$ different components of node embeddings that can represent the information of graph structure, node features, and features propagated through structure.

\emphasize{\textit{C1: Structure Embedding.}}
The structure embedding $\mathbf{U}$ is the left singular vector of the adjacency matrix $\mathbf{A}$, which is extracted by the singular value decomposition (SVD).
This aims to capture the community information of the graph.

\emphasize{\textit{C2: Neighborhood Embedding.}}
The neighborhood embedding $\mathbf{R}$ aims to capture the local higher-order neighborhood information of nodes.
By mimicking Personalized PageRank (PPR), we construct a random walk matrix $\mathbf{A}_{\text{PPR}}$, where each element is the number of times that a node visits another node in $T$ trials of the $k_{\text{PPR}}$-step random walks.
By doing random walks, the local higher-order structures will be highlighted among the entire graph.
To make $\mathbf{A}_{\text{PPR}}$ sparser and to speed up the embedding extraction, we eliminate the noisy elements with only one visited time.
We extract the left singular vectors of $\mathbf{A}_{\text{PPR}}$ by SVD as the neighborhood embeddings $\mathbf{R}$.

\emphasize{\textit{C3: Feature Embedding.}}
Given the raw node features $\mathbf{X}$, we represent the feature embedding with the preprocessed node features $\mathbf{F}=g(\mathbf{X})$, where $g$ is the preprocessed function.

\emphasize{\textit{C4: Propagation Embedding without Self-loop.}}
We row-normalize the adjacency matrix into $\mathbf{A}_{\text{row}} = \mathbf{D}^{-1}\mathbf{A}$, where $\mathbf{D}$ is the diagonal degree matrix.
The features are propagated without self-loop to capture the information of $k_{\text{row}}$-step neighbors, where $k_{\text{row}}$ is an even number.
This is useful to capture the information of similar neighbors when the structure exhibits heterophily (e.g., in a bipartite graph).
Therefore, we have node embedding $\mathbf{P}=g(l(\mathbf{A}_{\text{row}}^{2}\mathbf{X}))$, where $l$ is the column-wise L$2$-normalization, ensuring every dimension has a similar scale.

\emphasize{\textit{C5: Propagation Embedding with Self-loop.}}
The adjacency matrix with self-loop has been found useful to propagate the features in graphs that exhibit homophily.
Following the most common strategy, we symmetrically normalize the adjacency matrix into $\tilde{\mathbf{A}}_{\text{sym}} = (\mathbf{D} + \mathbf{I})^{-\frac{1}{2}}(\mathbf{A} + \mathbf{I})(\mathbf{D} + \mathbf{I})^{-\frac{1}{2}}$, where $\mathbf{I}$ is the identity matrix.
Similar to C$4$, we have node embeddings $\mathbf{S}=g(l(\tilde{\mathbf{A}}_{\text{sym}}^{k_{\text{sym}}}\mathbf{X}))$.

While C$1$-$2$ aim to capture the information with only the graph structure, C$4$-$5$ aim to capture the information of propagation, which is similar to the one that a trained GNN can capture.
To ensure that the embeddings have intuitive meanings, we set all the number of steps $k_{\text{PPR}}$, $k_{\text{row}}$ and $k_{\text{sym}}$ as $2$, which works sufficiently well in most cases.
As C$1$-$2$ adopted SVD as their last step, the embedding dimensions are orthogonalized.
For C$3$-$5$, we use principal component analysis (PCA) as $g$ to reduce and orthogonalize the embedding dimensions, leading to faster convergence and better performance when training the model.
Each component has the same number of dimensions $d$.

\subsection{\score: Definition and Theorems}
Next, we want to find a formula that connects the metrics of graph information and task performance.
To begin, we derive the inequality between entropy and accuracy:
\begin{theorem}[Entropy and Accuracy]
    Given a discrete random variable $Y$, we have:
    \begin{equation}
        2^{-H(Y)} \leq \text{accuracy}{(Y)} = \max_{y \in Y}{p_{y}}
    \end{equation}
    where $H(Y) = -\sum_{y \in Y}{p_{y} \log{p_{y}}}$ denotes the Shannon entropy.
    \label{thm:t1}
\end{theorem}
\begin{proof}
See Appx.~\ref{app:proof1}.
\end{proof}

Before extending Thm.~\ref{thm:t1} to the case with two random variables, we need a definition:
\begin{definition}[\score of Y given X] \label{def:score}
Given two discrete random variables $X$ and $Y$, \score of $Y$ given $X$ is defined as:
    \begin{equation} \label{eqn:score}
        \score = 2^{-H(Y|X)}
    \end{equation}
    where $H(\cdot | \cdot)$ denotes the conditional entropy.
\end{definition}

We prove that \score low-bounds the accuracy:
\begin{theorem}[\score]
    Given two discrete random variables $X$ and $Y$,
    \score of $Y$ given $X$ low-bounds the accuracy:
    \begin{equation}
        \score = 2^{-H(Y|X)} \leq \text{accuracy}{(Y|X)} = \sum_{x \in X}{\max_{y \in Y}{p_{x,y}}}
    \end{equation}
    where $p_{x,y}$ is the joint probability of $x$ and $y$.
    \label{thm:t2}
\end{theorem}
\begin{proof}
See Appx.~\ref{app:proof2}.
\end{proof}

\begin{figure}[t]
\begin{minipage}[t] {0.475\linewidth}
    \centering
    \subfloat[Link Prediction]
    {\includegraphics[height=1.27in]{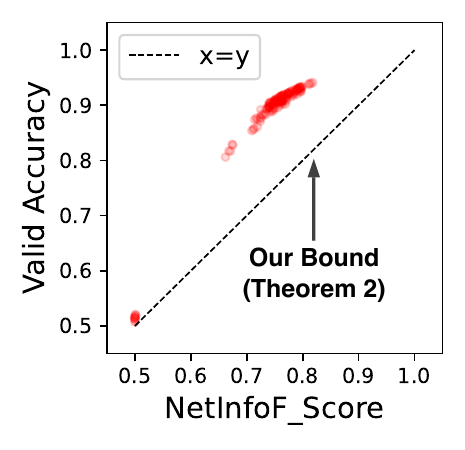}}
    \subfloat[Node Classification]
    {\includegraphics[height=1.27in]{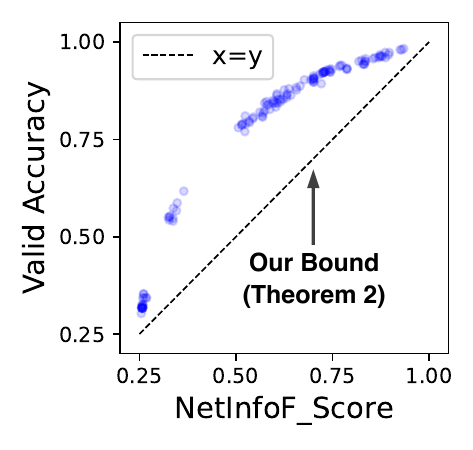}} 
    \caption{\emphasize{Thm.~\ref{thm:t2} holds}. \score is always less than or equal to validation accuracy. \label{fig:accnis}}
\end{minipage} \hfill
\begin{minipage}[t] {0.51\linewidth}
    \centering
    \subfloat[Link Prediction \label{fig:accnis_syn_lp}]
    {\includegraphics[height=1.27in]{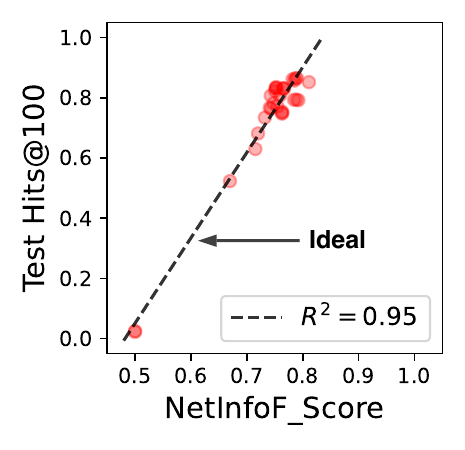}}
    \subfloat[Node Classification \label{fig:accnis_syn_nc}]
    {\includegraphics[height=1.27in]{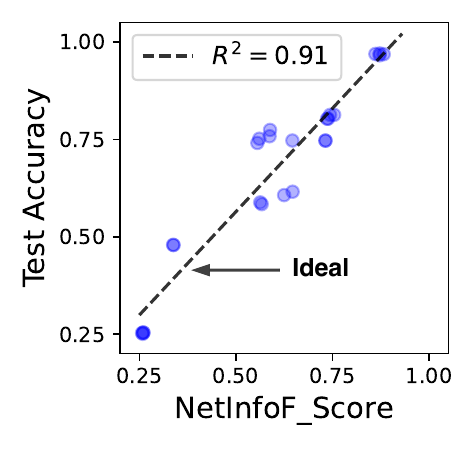}} 
    \caption{\emphasize{\score predicts right.} It is correlated to test performance in synthetic datasets. \label{fig:accnis_syn}}
\end{minipage}
\vspace{-2mm}
\end{figure}

% Although exactly computing the conditional entropy is difficult when there are many dimensions (Sec.~\ref{sec:related}), it is not necessary to do so.
Thm.~\ref{thm:t2} provides an advantage to \score by giving it an intuitive interpretation, which is the lower-bound of the accuracy.
When there is little usable information to the task, the value of \score is close to random guessing.
To empirically verify it, we run the experiments on the synthetic datasets (Appx.~\ref{app:data}) with five splits, and report \score and accuracy for all components of the derived embeddings.
In Fig.~\ref{fig:accnis}, we find that even for the validation set, \score is always less than or equal to the accuracy, strictly following Thm.~\ref{thm:t2}.
In the next sections, we show how \score can be effectively and efficiently computed with our proposed \analysis.

\section{\method for Link Prediction}
% Linear GNNs have shown to perform well on node classification.
% However, extending linear GNNs to link prediction is not easy, as node similarity plays an important role in this task.
% \textbf{Why node similarity is important in link prediction?}
% As discussed in \citet{wang2022flashlight}, the most outstanding link predictor of GNNs is a binary classifier $f$ (e.g., MLP or LogitReg) followed by the Hadamard product between the embeddings of the given node pair, i.e. $f(x_{i} \odot x_{j})$, other than the dot product $x_{i} \cdot x_{j}$ or the concatenation $f(x_{i} \parallel x_{j})$, where $x_{i}$ denotes the embedding of node $i$.
% We further dissect their observation and find this implies the two essential compositions link prediction, namely node similarity and dimension importance.
% The dot product handles node similarity, but ignores the dimension importance; in contrast, the concatenation handles the dimension importance, but ignores the node similarity.
% It is shown that the best link predictor of GNNs is a binary classifier $f$ (e.g., MLP) followed by the Hadamard product between the embeddings of the given node pair, i.e., $f(x_{i} \odot x_{j})$.
% Unlike non-linear GNNs, linear GNNs have static node embeddings, which limit them from:
% (1) precisely measuring node similarity when the neighbors have heterophily embeddings, and 
% (2) clearly distinguishing between the positive and negative edges.
% Therefore, in order to perform well in link prediction, a better node similarity measurement for linear GNNs is needed.

With the derived node embeddings, how can we measure NUI in link prediction as well as solve the task?
Compared to general GNNs, the node embeddings of linear GNNs are given by closed-form formula.
% Compared to general GNNs, the node embeddings of linear GNNs require no model training, but are static.
They are thus rarely applied on link prediction because of following two reasons:
(1) Predicting links by GNNs relies on measuring node similarity, which is incorrect if the neighbors are expected to have dissimilar embeddings;
% (1) Predicting links by GNNs relies on measuring node similarity, which is incorrect if the neighbors have heterophily embeddings.
for example, in a bipartite graph, while a source node is connected to a target node, their structural embeddings are expected to be very different, resulting in low node similarity by linear GNNs;
(2) In order to perform well on Hits@$K$, it is crucial to suppress the similarity of the nodes of negative edges, i.e. the unexisting connections in the graph.
Hits@$K$ is the ratio of positive edges that are ranked at $K$-th place or above among both the positive and negative edges, which is preferred in link prediction where most real-world applications are recommendations.
Since the embeddings of linear GNNs are static, they can not learn to separate the embeddings of nodes on each side of the negative edges.
% Compared to non-linear GNNs, linear GNNs are in an inferior position when doing link prediction as their derived embeddings are static.
% no learnable parameters to separate the embeddings of the positive edges from the ones of the negative edges.
Therefore, how to generalize linear GNNs to solve link prediction remains a challenge.
% \textcolor{red}{Visualization for the challenges?}

For these reasons, we propose an adjustment to the similarity of the nodes, which generalizes \method to link prediction, including \analysis to measure NUI and \model to solve the task.

\subsection{Proposed Adjustment to Node Similarity}
To solve the limitations of linear GNNs on link prediction, it is crucial to properly measure the similarity between nodes.
We consider cosine similarity as the measurement, whose value is normalized between $0$ and $1$.
By L$2$-normalizing each node embedding $\mathbf{z}_{1 \times d}$, the cosine similarity reduces to a simple dot product $\mathbf{z}_{i} \cdot \mathbf{z}_{j}$.
However, even if node $i$ and node $j$ are connected by an edge, it may result in low value if they are expected to have dissimilar embeddings (e.g. structure embeddings in a bipartite graph).
% It results in low value when node $i$ and node $j$ have heterophily embeddings, even if they are connected by an edge.
Therefore, before the dot product, we propose using the compatibility matrix $\mathbf{H}_{d \times d}$ to transform one of the embeddings, and rewrite the node similarity function into $\mathbf{z}_{i} \mathbf{H} \mathbf{z}_{j}^{\intercal}$.\looseness=-1

The compatibility matrix $\mathbf{H}$ represents the characteristics of the graph: if the graph exhibits homophily, $\mathbf{H}$ is nearly diagonal; if it exhibits heterophily, $\mathbf{H}$ is off-diagonal. 
It is commonly assumed, given in belief propagation (BP) to handle the interrelations between node classes.
In our case, $\mathbf{H}$ represents the interrelations between the dimensions of the node embeddings.
% However, it can not be estimated by the solution in BP because the embeddings are continuous, and even if we discretize the features, each node will be multi-class.
By maximizing the similarity of nodes connected by edges, $\mathbf{H}$ can be estimated by the following lemma:
\begin{lemma}[Compatibility Matrix]\label{lem:cheapcm}
The compatibility matrix $\mathbf{H}$ has the closed-form solution and can be solved by the following optimization problem:
\begin{equation}
    \min_{\mathbf{H}}{\sum_{(i, j) \in \mathcal{E}}{\| \mathbf{z}_{i}\mathbf{H} - \mathbf{z}_{j} \|^{2}_{2}}},
\end{equation}
where $\mathcal{E}$ denotes the set of (positive) edges in the given graph.
\end{lemma}
\begin{proof}
See Appx.~\ref{app:proof3}.
\end{proof}
This optimization problem can be efficiently solved by multi-target linear regression.
Nevertheless, this estimation of $\mathbf{H}$ does not take into account negative edges, which may accidentally increase the similarity of negative edges in some complicated cases.
This hurts the performance especially when evaluating with Hits@$K$.
Therefore, based on Lemma~\ref{lem:cheapcm}, we propose an improved estimation $\mathbf{H}^{*}$, which further minimizes the similarity of nodes connected by the sampled negative edges:
\begin{lemma}[Compatibility Matrix with Negative Edges]\label{lem:expcm}
The compatibility matrix with negative edges $\mathbf{H}^{*}$ has the closed-form solution and can be solved by the following optimization problem:
\begin{equation}
    \min_{\mathbf{H}^{*}}{\sum_{(i, j) \in \mathcal{E}}{(1- \mathbf{z}_{i}\mathbf{H}^{*}\mathbf{z}_{j}^{\intercal}})-\sum_{(i, j)\in \mathcal{E}_{\text{neg}}}{(\mathbf{z}_{i}\mathbf{H}^{*}\mathbf{z}_{j}^{\intercal})}},
\end{equation}
where $\mathcal{E}_{\text{neg}}$ denotes the set of negative edges.
\end{lemma}
\begin{proof}
See Appx.~\ref{app:proof4}.
\end{proof}

With great power comes great responsibility, estimating $\mathbf{H}^{*}$ has a higher computational cost than estimating $\mathbf{H}$.
Thus, we provide three techniques to speed up the computation of $\mathbf{H}^{*}$ with the help of $\mathbf{H}$, and the details are in Algo.~\ref{algo:a1}.

\emphasize{\textit{T1: Warm Start.}}
We approximate the solution by LSQR iteratively and warm up the approximation process with $\mathbf{H}$.
Since $\mathbf{H}$ is similar to $\mathbf{H}^{*}$ and cheap to compute, this step largely speeds up the approximation process and reduces the number of iterations needed for convergence.

\emphasize{\textit{T2: Coefficient Selection.}}
We reduce the number of coefficients by only estimating the upper triangle of $\mathbf{H}^{*}$, and keep the ones with $95\%$ energy in $\mathbf{H}$.
This is because the similarity function is symmetric, and the unimportant coefficients with small absolute values in $\mathbf{H}$ remain unimportant in $\mathbf{H}^{*}$.
The absolute sum of the kept coefficients divided by $\sum_{i=1}^{d}{\sum_{j=i+1}^{d}{|\mathbf{H}_{ij}|}}$ is $95\%$ and the rest are zeroed out.
This helps us reduce the number of coefficients from $d^{2}$ to be less than $(d+1)d/2$.

\emphasize{\textit{T3: Edge Reduction.}}
We sample $S$ positive edges from the $2$-core graph, and $2S$ negative edges, where the sample size $S$ depends on $d$.
Since in large graphs $|\mathcal{E}|$ is usually much larger than $d^{2}$, it is not necessary to estimate fewer than $(d+1)d/2$ coefficients with all $|\mathcal{E}|$ edges.
Moreover, the $2$-core graph remains the edges with stronger connections, where each node in it has at least degree $2$.
Sampling from the $2$-core graph avoids interference from noisy edges and leads to better estimation.
% In our experiment, where $d=128$, we at most use $200$K positive and $400$K negative edges, and still get very good performance empirically.

\subsection{\analysis for NUI Measurement}
Based on Thm.~\ref{thm:t2}, we propose \analysis that computes \score, without exactly computing the conditional entropy of the high-dimensional variables.
By sampling negative edges, the link prediction can be seen as a binary classification problem.
For each component of embeddings, \analysis esitmates its corresponding $\mathbf{H}^{*}$ and discretizes the adjusted node similarity of positive and negative edges.
To avoid overfitting, we fit the $k$-bins discretizer with the similarity of training edges, and discretize the one of validation edges into $k$ bins.
\score can then be easily computed between two categorical variables.
For instance, the node similarity between node $i$ and $j$ with embedding $\mathbf{U}$ is $(\hat{\mathbf{U}}_{i}\mathbf{H}^{*}_{\hat{\mathbf{U}}}) \cdot \hat{\mathbf{U}}_{j}$, where $\hat{\cdot}$ denotes the embedding preprocessed by column-wise standardization and row-wise L$2$-normalization.
The details are in Algo.~\ref{algo:a2}.

\subsection{\model for NUI Exploitation}
To solve link prediction, \model shares the same derived node embeddings with \analysis, and uses a link predictor following by the Hadamard product of the embeddings.
We transform the embeddings on one side of the edge with $\mathbf{H}^{*}$, which handles the heterophily embeddings and better separates the nodes in the negative edges.
By concatenating all components, the input to the predictor is as follows:
\begin{equation}
\underbrace{\hat{\mathbf{U}}_{i}\mathbf{H}^{*}_{\hat{\mathbf{U}}} \odot \hat{\mathbf{U}}_{j}}_{\text{Structure}} \cat
\underbrace{\hat{\mathbf{R}}_{i}\mathbf{H}^{*}_{\hat{\mathbf{R}}} \odot \hat{\mathbf{R}}_{j}}_{\text{PPR}} \cat
\underbrace{\hat{\mathbf{F}}_{i}\mathbf{H}^{*}_{\hat{\mathbf{F}}} \odot \hat{\mathbf{F}}_{j}}_{\text{Features}} \cat
\underbrace{\hat{\mathbf{P}}_{i}\mathbf{H}^{*}_{\hat{\mathbf{P}}} \odot \hat{\mathbf{P}}_{j}}_{\substack{\text{Features of} \\ \text{$2$-Step Neighbors}}} \cat
\underbrace{\hat{\mathbf{S}}_{i}\mathbf{H}^{*}_{\hat{\mathbf{S}}} \odot \hat{\mathbf{S}}_{j}}_{\substack{\text{Features of} \\ \text{Grand Neighbors}}}
\end{equation}
where $(i, j) \in \mathcal{E} \cup \mathcal{E}_{\text{neg}}$.
Among all the choices, we use LogitReg as the predictor for its scalability and interpretability.
We suppress the weights of useless components, if there is any, by adopting sparse-group LASSO for the feature selection.
The time complexity of \model is:
\begin{lemma} \label{lem:time}
The time complexity of \model for link prediction is linear on the input size
$|\mathcal{E}|$:\looseness=-1
\begin{equation}
O(f^{2}|\mathcal{V}| + f^{3} + d^{4}|\mathcal{E}|)
\end{equation}
where $f$ and $d$ are the number of features and embedding dimensions, respectively.
\end{lemma}
\begin{proof}
See Appx.~\ref{app:proof5}.
\end{proof}

\label{sec:tasklp}

\section{\method for Node Classification}
In this section, we show how we can generalize \method to node classification.
% We evaluate the performance with the synthetic datasets for sanity checks in Appx.~\ref{app:nc_syn}.
In contrast to link prediction, node classification does not rely on the node similarity, needing no compatibility matrix.

\subsection{\analysis for NUI Measurement}
% The exact computation of conditional entropy for \score is prohibitive.
To effectively and efficiently compute \score, we propose to assign labels to the nodes by clustering.
This idea is based on the intuition that good embeddings for node classification can be easily split by clustering.
Among clustering methods, we use $k$-means as it is fast.
We cluster each component of the embeddings and compute \score, where $k \geq c$.
To ensure that the clustering is done stably, a row-wise L$2$-normalization is done on the embedding. 
The details are in Algo.~\ref{algo:a3}.

\subsection{\model for NUI Exploitation}
To solve node classification, we again concatenate the embeddings of different components, and the input of classifier is as follows:
\begin{equation}
\underbrace{l(\mathbf{U})}_{\text{Structure}} \cat
\underbrace{l(\mathbf{R})}_{\text{PPR}} \cat
\underbrace{l(\mathbf{F})}_{\text{Features}} \cat
\underbrace{l(\mathbf{P})}_{\substack{\text{Features of} \\ \text{$2$-Step Neighbors}}} \cat
\underbrace{l(\mathbf{S})}_{\substack{\text{Features of} \\ \text{Grand Neighbors}}}
\end{equation}
where $l$ is the column-wise L$2$-normalization.
Similar to \model in link prediction, we use LogitReg as the classifier and adopt sparse-group LASSO for the regularization.

\section{Synthetic Datasets for Sanity Checks}
To ensure that \method is robust to all graph scenarios, we carefully design the synthetic datasets for sanity checks.
We include all possible graph scenarios, where the ground truth of NUI is available.
The details of implementation are in Appx.~\ref{app:data}.

% To check the correctness of \analysis and the robustness of \model, they are evaluated on our carefully designed synthetic datasets, whose ground truth NUI is available, with all possible graph scenarios.

\subsection{Link Prediction}
\begin{figure}[t]
\begin{minipage}{0.43\textwidth}
\centering
\includegraphics[scale=0.44]{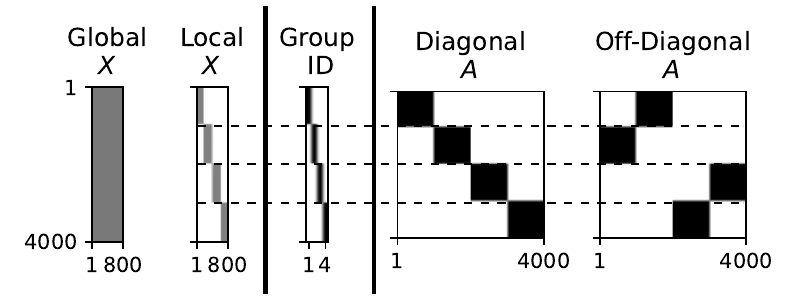}
\captionof{figure}{Scenarios of node features $\mathbf{X}$ and graph structure $\mathbf{A}$ in synthetic datasets.}
\label{fig:synthetic}
\end{minipage}
\hfill
\begin{minipage}{0.56\textwidth}
\captionof{table}{
    \emphasize{\method wins} on link prediction in the synthetic datasets. 
    Hits@$100$ is reported.
}
\centering{\resizebox{1\textwidth}{!}{
\begin{tabular}{ l | cc | cc | cc | r }
	\toprule
	\multirow{2}{*}{\textbf{Model}}
	    & \textbf{Rand. $\mathbf{X}$}
	    & \textbf{Rand. $\mathbf{X}$} 
	    & \textbf{Global $\mathbf{X}$} 
        & \textbf{Global $\mathbf{X}$}
	    & \textbf{Local $\mathbf{X}$}
	    & \textbf{Local $\mathbf{X}$}
        & \textbf{Avg.} \\
	    & \textbf{Diag. $\mathbf{A}$}
        & \textbf{Off-Diag. $\mathbf{A}$}
	    & \textbf{Diag. $\mathbf{A}$} 
	    & \textbf{Off-Diag. $\mathbf{A}$} 
	    & \textbf{Diag. $\mathbf{A}$}
	    & \textbf{Off-Diag.}
        & \textbf{Rank} \\
	\midrule
	GCN
	  & 82.7$\pm$1.1
        & 70.9$\pm$1.2
	    & 87.2$\pm$0.5
	    & 85.1$\pm$1.2
	    & 17.4$\pm$1.7
	    & 19.2$\pm$1.8
	    & 4.3 (0.9) \\
	SAGE
	    & 77.4$\pm$1.1
        & 66.5$\pm$4.1
	    & 86.2$\pm$1.0
	    & 85.2$\pm$2.2
	    & 11.0$\pm$1.2
	    & 09.5$\pm$1.0
	    & 5.1 (1.1) \\
	GAT
	    & 86.3$\pm$0.9
        & 83.1$\pm$0.3
	    & 87.3$\pm$0.9
	    & 85.2$\pm$0.6
	    & 16.0$\pm$2.0
	    & 16.9$\pm$2.1
	    & 3.3 (1.5) \\
    H$^{2}$GCN
	    & 24.5$\pm$4.3
	    & 58.9$\pm$3.7
        & 75.3$\pm$1.1
	    & 85.8$\pm$2.5
	    & 19.8$\pm$2.0
	    & 19.2$\pm$1.5
	    & 5.0 (2.1) \\
	GPR-GNN
	    & 75.1$\pm$0.8
	    & 52.3$\pm$1.6
	    & 83.4$\pm$1.3
	    & 79.5$\pm$1.6
        & 19.3$\pm$1.7
	    & 17.1$\pm$2.0
	    & 6.0 (1.1) \\
    \textsc{SlimG}
	    & 85.7$\pm$0.8
	    & 67.8$\pm$2.8
	    & 87.9$\pm$1.0
        & 85.1$\pm$1.3
	    & 82.5$\pm$1.6
	    & 31.1$\pm$1.1
	    & 3.3 (1.3) \\
    \midrule
	\method
	    & \textbf{87.3$\pm$0.7}
	    & \textbf{86.7$\pm$0.6}
	    & \textbf{89.8$\pm$0.3}
        & \textbf{89.8$\pm$1.0}
	    & \textbf{89.6$\pm$0.2}
	    & \textbf{90.8$\pm$0.7}
	    & \textbf{1.0 (0.0)} \\
	\bottomrule
\end{tabular}
}}
\label{table:exprlp_syn}
\end{minipage}
\end{figure}

\textbf{Designs.}
We separate the nodes into $c$ groups to simulate that there are usually multiple communities in a graph.
To cover all the possibilities in the real-world, the scenarios are the cross-product of different scenarios on the node features $\mathbf{X}$ and the graph structure $\mathbf{A}$, as shown in Fig.~\ref{fig:synthetic}.
We ignore the scenario that $\mathbf{X}$ is useful but $\mathbf{A}$ is useless, since this is impractical in the real-world.

There are $3$ scenarios of node features $\mathbf{X}$:
\begin{compactenum}
\item \textbf{Random}: the node features are random, with no correlation with the existence of edges.
\item \textbf{Global}: all dimensions of the node features are correlated with the existence of edges.
\item \textbf{Local}: only a subset of dimensions of the node features are correlated with the existence of edges, where there is no overlapping between the subsets of node groups.
\end{compactenum}
There are $2$ scenarios of graph structure $\mathbf{A}$:
\begin{compactenum}
\item \textbf{Diagonal}: the nodes and their neighbors are in the same group.
\item \textbf{Off-Diagonal}: the nodes and their neighbors are in two different groups.
\end{compactenum}

\textbf{Observations.}
In Table~\ref{table:exprlp_syn}, \method receives the highest average rank among all GNN baselines, and is the only method that can handle all scenarios.
While GNNs have worse performance when $\mathbf{X}$ is either random or local, \textsc{SlimG}, a linear GNN, cannot handle cases with off-diagonal $\mathbf{A}$.

\textbf{Would a GNN work?}
Fig.~\ref{fig:accnis_syn_lp} shows that \score is highly correlated with test Hits@$100$, with high $R^{2}$ values, where each point denotes a component of embeddings from each split of synthetic datasets.
In Appx.~\ref{app:lp_sc}, Table~\ref{table:exprlp_syn_info} reports \score and test performance of each component.
By measuring their \score, \analysis tells when propagating features through structure contains less information than using features or structure itself.
For example, in scenarios where node features are useless (the first two scenarios in Table~\ref{table:exprlp_syn_info}), \analysis spots that $\mathbf{F}$ (i.e., $g(\mathbf{X})$) provides little NUI to the task, and thus the propagated embeddings $\mathbf{P}$ and $\mathbf{S}$ have less NUI than the structural embeddings $\mathbf{U}$ and $\mathbf{R}$. 
This indicates that training GNNs is less likely to have a better performance than only utilizing the information from the graph structure, which correctly matches the test performance in Table~\ref{table:exprlp_syn_info}.
% \score is as well helpful for the users to analyze the given graph, which spots the useless information that should not be included for the further usage.

\subsection{Node Classification}

\textbf{Designs.}
We remain the same scenarios in \textsc{SlimG}, while using our graph generator in Appx.~\ref{app:data}.

\textbf{Observations.}
Fig.~\ref{fig:accnis_syn_nc} shows that \score is highly correlated with test accuracy.
In Appx.~\ref{app:nc_syn}, Table~\ref{table:exprnc_syn} shows that \method generalizes to all scenarios as \textsc{SlimG} does; Table~\ref{table:exprnc_syn_info} shows that the component with the highest \score always has the highest test accuracy.

\section{Experiments}
% \textcolor{blue}{We do not answer the question: Given a graph with node features, how to tell if a graph neural network (GNN) can perform well on graph tasks or not? However this question is the motivation of NetInfo}

We conduct experiments by real-world graphs to answer the following research questions (RQ):
\begin{compactenum}[{RQ}1.]
    \item {\bf Effectiveness:} How well does \method perform in real-world graphs?
    \item {\bf Scalability:} Does \method scales linearly with the input size?
    \item {\bf Ablation Study:} Are all the design choices in \method necessary?
\end{compactenum}
The details of datasets and settings are in Appx.~\ref{app:rep}.
Since we focus on improving linear GNNs in link prediction, the experiments for node classification are in Appx.~\ref{app:nc_real} because of space limit.
The experiments are conducted on an AWS EC2 G4dn instance with $192$GB RAM.

\subsection{Effectiveness (RQ1)}
\paragraph{Real-World Datasets.}
We evaluate \method on $7$ homophily and $5$ heterophily real-world graphs.
We randomly split the edges into training, validation, and testing sets by the ratio $70\%/10\%/20\%$ five times and report the average for fair comparison.
Since our goal is to propose a general GNN method, we focus on comparing \method with $6$ GNN baselines, which are general GNNs (GCN, SAGE, GAT), heterophily GNNs (H$^{2}$GCN, GPR-GNN), and a linear GNN (\textsc{SlimG}).
While Hits@$100$ is used for evaluating on most graphs, Hits@$1000$ is used on the larger ones, namely, Products, Twitch, and Pokec, which have much more negative edges in the testing sets.\looseness=-1

In Table~\ref{table:exprlp}, \method outperforms GNN baselines in $11$ out of $12$ datasets, and has the highest average rank, as our derived embeddings include comprehensive graph information, that is, structure, features, and features propagated through structure.
Compared to non-linear GNNs, \textsc{SlimG} performs worse in most heterophily graphs, showing that it cannot properly measure the node similarity of heterophily embeddings in link prediction.
By addressing the limitations of linear GNNs, \method is able to consistently outperform both \textsc{SlimG} and non-linear GNNs in both homophily and heterophily graphs.
Note that the results in Pokec are similar to the ones in homophily graphs, since it can be labeled as either homophily (by locality) or heterophily (by gender).

\begin{table*}[h]
\caption{
    \emphasize{\method wins} on link prediction in most real-world datasets. 
    Hits@$100$ is reported for most datasets, and Hits@$1000$ for the large datasets (Products, Twitch, and Pokec).
}
\centering{\resizebox{\textwidth}{!}{
\begin{tabular}{ l | ccccccc | ccccc | r }
	\toprule
	\textbf{Model}
	    & \textbf{Cora}
	    & \textbf{CiteSeer} 
	    & \textbf{PubMed} 
	    & \textbf{Comp.}
	    & \textbf{Photo}
	    & \textbf{ArXiv}
	    & \textbf{Products}
	    & \textbf{Cham.} 
	    & \textbf{Squirrel} 
	    & \textbf{Actor} 
	    & \textbf{Twitch}
	    & \textbf{Pokec}
        & \textbf{Avg. Rank} \\
	\midrule
	GCN
	  & 67.1$\pm$1.8
	    & 60.4$\pm$10.
	    & 47.6$\pm$13.
	    & 22.5$\pm$3.1
	    & 39.1$\pm$1.6
	    & 14.8$\pm$0.6
	    & 02.2$\pm$0.1
	    & 82.1$\pm$4.5
	    & 16.5$\pm$1.0
	    & 31.1$\pm$1.7
	    & 16.2$\pm$0.3
	    & 07.9$\pm$1.7
	    & 4.1 (1.3) \\
	SAGE
	    & 68.4$\pm$2.8
	    & 55.9$\pm$2.5
	    & 57.6$\pm$1.1
	    & 27.5$\pm$2.1
	    & 40.0$\pm$1.9
	    & 00.7$\pm$0.1
	    & 00.3$\pm$0.2
	    & 84.7$\pm$3.6
	    & 15.5$\pm$1.5
	    & 27.6$\pm$1.4
	    & 08.7$\pm$0.6
	    & 05.5$\pm$0.5
	    & 4.5 (1.2) \\
	GAT
	    & 66.7$\pm$3.6
	    & 65.2$\pm$2.6
	    & 55.1$\pm$2.4
	    & 28.3$\pm$1.6
	    & 44.2$\pm$3.5
	    & 05.0$\pm$0.8
	    & O.O.M.
	    & 84.8$\pm$4.5
	    & 15.6$\pm$0.8
	    & 32.3$\pm$2.4
	    & 08.2$\pm$0.3
	    & O.O.M.
	    & 4.0 (1.7) \\
    H$^{2}$GCN
	    & 64.4$\pm$3.4
	    & 35.7$\pm$5.4
	    & 50.5$\pm$0.9
	    & 17.9$\pm$0.7
	    & 29.5$\pm$2.4
	    & O.O.M.
        & O.O.M.
	    & 79.3$\pm$4.5
	    & 16.0$\pm$2.6
	    & 28.7$\pm$2.1
	    & O.O.M.
	    & O.O.M.
	    & 6.2 (1.0) \\
	GPR-GNN
	    & 69.8$\pm$1.9
	    & 53.5$\pm$8.1
	    & \textbf{66.3$\pm$3.3}
	    & 20.7$\pm$1.8
	    & 34.1$\pm$1.1
	    & 13.8$\pm$0.8
	    & O.O.M.
	    & 77.2$\pm$5.6
	    & 14.6$\pm$2.7
	    & 32.1$\pm$1.3
	    & 12.6$\pm$0.2
	    & 05.0$\pm$0.2
	    & 4.6 (1.8) \\
    \textsc{SlimG}
	    & 77.9$\pm$1.3
	    & 86.8$\pm$1.0
	    & 55.9$\pm$2.8
	    & 25.3$\pm$0.9
	    & 40.2$\pm$2.5
	    & 20.2$\pm$1.0
	    & 27.6$\pm$0.6
	    & 76.9$\pm$2.8
	    & 19.6$\pm$1.5
	    & 18.7$\pm$1.0
	    & 12.0$\pm$0.3
	    & 21.7$\pm$0.2
	    & 3.5 (1.8) \\
	\midrule
	\method
	    & \textbf{81.3$\pm$0.6}
	    & \textbf{87.3$\pm$1.3}
	    & 59.7$\pm$1.1
	    & \textbf{31.1$\pm$1.9}
	    & \textbf{46.8$\pm$2.2}
	    & \textbf{39.2$\pm$1.8}
	    & \textbf{35.2$\pm$1.1}
	    & \textbf{86.9$\pm$2.3}
	    & \textbf{24.2$\pm$2.0}
	    & \textbf{36.2$\pm$1.2}
	    & \textbf{19.6$\pm$0.7}
	    & \textbf{31.3$\pm$0.5}
	    & \textbf{1.1 (1.3)} \\
	\bottomrule
\end{tabular}
}}
\label{table:exprlp}
\end{table*}

\clearpage
\paragraph{OGB Link Prediction Datasets.}
We evaluate \method on OGB datasets.
Table~\ref{table:ogbl} shows that \method outperforms other general GNN baselines, while using a model with much fewer parameters.
\method has only $1280$ learnable parameters for all datasets, while GCN and SAGE have at least $279$K and $424$K, respectively, which is $218\times$ more than the ones that \method has.

\subsection{Scalability (RQ2)}
\begin{figure}[t]
\begin{minipage}{0.47\textwidth}
\centering
\includegraphics[scale=0.5]{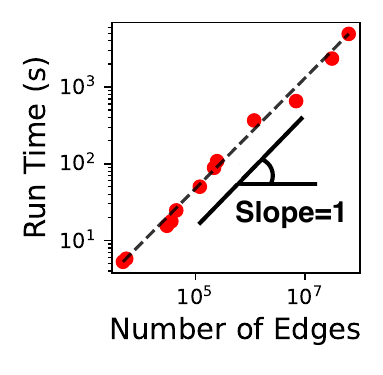}
\vspace{-5mm}
\captionof{figure}{\emphasize{\model is scalable} on link prediction, being linear with number of edges.}
\label{fig:scalable}
\end{minipage}
\hfill
\begin{minipage}{0.51\textwidth}
\captionof{table}{
\emphasize{\method wins} on link prediction in OGB datasets.
Hits@$20$, Hits@$50$, and Hits@$100$ are reported for ddi, collab, and ppa, respectively.
}
\centering{\resizebox{0.8\textwidth}{!}{
\begin{tabular}{ l | ccc }
	\toprule
	\textbf{Model}
	    & \textbf{ogbl-ddi}
	    & \textbf{ogbl-collab} 
	    & \textbf{ogbl-ppa} \\
	\midrule
	GCN
	  & 37.1$\pm$5.1
	    & 44.8$\pm$1.1
	    & 18.7$\pm$1.3 \\
    SAGE
	  & 53.9$\pm$4.7
	    & 48.1$\pm$0.8
	    & 16.6$\pm$2.4 \\
    \textsc{SlimG}
	  & 35.9$\pm$0.6
	    & 45.1$\pm$0.2
	    & 21.1$\pm$0.6 \\
	\midrule
	\method
	    & \textbf{56.8$\pm$3.4}
	    & \textbf{53.7$\pm$0.2}
	    & \textbf{24.2$\pm$0.1} \\
	\bottomrule
\end{tabular}
}}
\label{table:ogbl}
\end{minipage}
\end{figure}

We plot the number of edges versus the run time of link prediction in seconds on the real-world datasets.
In Fig.~\ref{fig:scalable}, we find that \method scales linearly with the number of edges, thanks to our speed-up techniques in estimating compatibility matrix $\mathbf{H}^{*}$.
To give a concrete example, the numbers of coefficients of $\mathbf{H}^{*}_{\mathbf{U}}$ are reduced from $d(d+1)/2=8256$ to $3208$, $5373$, and $4293$, for Products, Twitch, and Pokec, respectively.
Moreover, those numbers are very reasonable: Products is a homophily graph, its $\mathbf{H}^{*}_{\mathbf{U}}$ has the fewest coefficients, which are mostly on the diagonal; Twitch is a heterophily graph, its $\mathbf{H}^{*}_{\mathbf{U}}$ has the most coefficients, which are mostly on the off-diagonal; Pokec can be seen as either homophily or heterophily, its $\mathbf{H}^{*}_{\mathbf{U}}$ has the number of coefficients in between.

\subsection{Ablation Study (RQ3)}
% In Sec.~\ref{sec:tasklp}, we point out two challenges of extending linear GNNs to link prediction.
To demonstrate the necessity of addressing the limitations of linear GNNs in link prediction with our design choices, we study \method (1) without compatibility matrix (w/o CM), and (2) with only compatibility matrix $\mathbf{H}$ (w/ only $\mathbf{H}$), which is not optimized with negative edges.
Table~\ref{table:ablation} shows that \method works best with both design choices.
In heterophily graphs, merely using $\mathbf{H}$ leads to better performance because of properly handling heterophily embeddings; while in homophily graphs, it accidentally increases the similarity between nodes in negative edges and hurts the performance.
Taking into account both heterophily embeddings and negative edges, using $\mathbf{H}^{*}$ as the compatibility matrix has the best performance in both heterophily and homophily graphs.

% However, it is also shown that even for the larger homophily graphs (namely Computers, ArXiv, and Products), using compatibility matrix still leads to better performance.
% This indicates that when the homophily graph becomes more complicated, the embedding may not stay homophily, which again emphasizes the necessarity of compatibility matrix.

\begin{table*}[h]
\caption{
    \emphasize{Ablation Study - All design choices in \method are necessary} on link prediction. 
    CM stands for compatibility matrix, and $\mathbf{H}$ is not optimized with negative edges.
    % Hits@$100$ is reported for most datasets, and Hits@$1000$ for the large datasets (Products, Twitch and Pokec).
}
\vspace{-1mm}
\centering{\resizebox{\textwidth}{!}{
\begin{tabular}{ l | ccccccc | ccccc }
	\toprule
	\textbf{Model}
	    & \textbf{Cora}
	    & \textbf{CiteSeer} 
	    & \textbf{PubMed} 
	    & \textbf{Comp.}
	    & \textbf{Photo}
	    & \textbf{ArXiv}
	    & \textbf{Products}
	    & \textbf{Cham.} 
	    & \textbf{Squirrel} 
	    & \textbf{Actor} 
	    & \textbf{Twitch}
	    & \textbf{Pokec} \\
	\midrule
	w/o CM
	  & 79.8$\pm$0.9
	    & 86.5$\pm$1.6
	    & 58.5$\pm$1.2
	    & 27.9$\pm$0.2
	    & 44.7$\pm$3.2
	    & 35.9$\pm$1.8
	    & 34.6$\pm$0.4
	    & 74.6$\pm$1.5
	    & 14.3$\pm$0.6
	    & 29.5$\pm$1.8
	    & 08.9$\pm$0.8
	    & 30.5$\pm$0.3 \\
    w/ only $\mathbf{H}$
	  & 80.9$\pm$0.5
	    & 87.0$\pm$1.4
	    & 58.8$\pm$1.4
	    & 26.5$\pm$1.2
	    & 43.4$\pm$2.1
	    & 32.5$\pm$1.6
	    & 30.6$\pm$0.4
	    & 74.3$\pm$3.2
	    & 19.3$\pm$1.2
	    & 32.2$\pm$1.6
	    & 10.3$\pm$3.1
	    & 29.8$\pm$0.4 \\
	\midrule
	\method
	    & \textbf{81.3$\pm$0.6}
	    & \textbf{87.3$\pm$1.3}
	    & \textbf{59.7$\pm$1.1}
	    & \textbf{31.1$\pm$1.9}
	    & \textbf{46.8$\pm$2.2}
	    & \textbf{39.2$\pm$1.8}
	    & \textbf{35.2$\pm$1.1}
	    & \textbf{86.9$\pm$2.3}
	    & \textbf{24.2$\pm$2.0}
	    & \textbf{36.2$\pm$1.2}
	    & \textbf{19.6$\pm$0.7}
	    & \textbf{31.3$\pm$0.5} \\
	\bottomrule
\end{tabular}
}}
\label{table:ablation}
\end{table*}

% \subsection{GNN Suitability (RQ3)}
% Most real-world graphs that used by previous studies usually have both usable node features and graph structure.
% Therefore, we use a heterogeneous shopping queries network ESCI \citep{reddy2022shopping}, which includes four types of edges, namely, ``Exact'', ``Substitute'', ``Complement'' and "Irrelevant".
% The intuition is that the graph constructed by irrelevant edges should have both useless features and useless structure.
% In Table~\ref{table:story}, the results of validation \score fit our intuition, where \score of ``Irrelevant'' edges are close to random guessing.
% The edges with types ``Substitute'' and ``Complement'' contain no usable information, and link prediction can not be done well on them.

\section{Conclusions}
We propose the \method framework to measure and exploit the network usable information (NUI).
% The first module \analysis measures NUI with \score, and the second module \model solves downstream graph tasks.
In summary, \method has the following advantages:
\ben
\item \general, handling both link prediction and node classification (Lemma~\ref{lem:cheapcm}-\ref{lem:expcm});
\item \principled, with theoretical guarantee (Thm.~\ref{thm:t1}-\ref{thm:t2}) and closed-form solution (Lemma~\ref{lem:cheapcm}-\ref{lem:expcm});
\item \effective, thanks to the proposed adjustment of node similarity;
\item \scalable, scaling linearly with the input size (Fig.~\ref{fig:scalable}).
\een
Applied on our carefully designed synthetic datasets, \method correctly identifies the ground truth of NUI and is the only method that is robust to all graph scenarios.
Applied on real-world graphs, \method wins in \textit{11 out of 12} times on link prediction.

\clearpage
\bibliography{BIB/ref}
\bibliographystyle{iclr2024_conference}

\clearpage
\appendix
\section{Appendix: Proofs}
\label{app:proofs}

\subsection{Proof of Thm.~\ref{thm:t1}} \label{app:proof1}
\begin{proof}
\label{proof:p1}
Let $Y$ be a discrete random variable with
$n$ outcomes ($y_1, \ldots, y_n$),
and with probabilities ($\pOne, \ldots, \pEn$), where:
\begin{equation}
    1 = \pOne + \ldots + \pEn
\end{equation}
Let $\pMax$ be the highest probability
(break ties arbitrarily), that is:
\begin{equation}
    \pMax = \max_i \pEye
\end{equation}
For ease of presentation, and without loss of generality,
assume that the most likely outcome is the first one, $y_1$, and thus $p_{max} = p_1$.
Given no other information, the best classifier 
for $Y$ is the one that always guesses outcome $y_1$, and it has accuracy:
\begin{equation}
    \text{accuracy}(Y) = \pOne \equiv \pMax
\end{equation}
The entropy $H(Y)$ is:
\begin{equation}
       H(Y) =  - (\pOne \log \pOne + \ldots + \pEn \log \pEn)
\end{equation}
Thus we have:
\begin{align}
        2^{- H(Y) } & =  \pOne ^ {\pOne} * \pTwo ^ {\pTwo} \ldots * \pEn ^ {\pEn} \\
           & \leq \pMax ^ {\pOne} * \pMax ^{\pTwo} * \ldots *
           \pMax ^ {\pEn} ~~~~// \ \because \pMax \geq \pEye \\
           & \leq \pMax ^ {\pOne + \pTwo + \ldots + \pEn} \\
           & \leq \pMax
\end{align}
which completes the proof.
\end{proof}

\subsection{Proof of Thm.~\ref{thm:t2}} \label{app:proof2}
\begin{proof}
Let $Y$ and $X$ be two discrete random variables with $n$ outcomes ($y_1, \ldots, y_n$) and $m$ outcomes ($x_1, \ldots, x_m$), respectively, then their joint probabilities are ($p_{1, 1}, \ldots, p_{m, n}$), where:
\begin{equation}
    1 = p_{1} + \ldots + p_{m} = \sum_{j=1}^{n}{p_{1, j}} + \ldots + \sum_{j=1}^{n}{p_{m, j}}
\end{equation}
The accuracy is:
\begin{equation}
    \text{accuracy}(Y | X) = \sum_{i=1}^{m}{\max_{j}{p_{i, j}}}
\end{equation}
The conditional entropy $H(Y | X)$ is:
\begin{align}
H(Y | X) & = p_{1} * (-\sum_{j=1}^{n}{\frac{p_{1, j}}{p_{1}} * \log_{2}{\frac{p_{1, j}}{p_{1}}}}) + \ldots + p_{m} * (-\sum_{j=1}^{n}{\frac{p_{m, j}}{p_{m}} * \log_{2}{\frac{p_{m, j}}{p_{m}}}})
\end{align}
Thus we have:
\begin{align}
\Rightarrow -H(Y | X) & = p_{1} * (\sum_{j=1}^{n}{\frac{p_{1, j}}{p_{1}} * \log_{2}{\frac{p_{1, j}}{p_{1}}}}) + \ldots + p_{m} * (\sum_{j=1}^{n}{\frac{p_{m, j}}{p_{m}} * \log_{2}{\frac{p_{m, j}}{p_{m}}}}) \\
& \leq p_{1} * \log_{2}{(\sum_{j=1}^{n}{(\frac{p_{1, j}}{p_{1}})^{2}})} + \ldots + p_{m} * \log_{2}{(\sum_{j=1}^{n}{(\frac{p_{m, j}}{p_{m}})^{2}})} ~~~~// \ \because \text{Jensen's Inequality} \\
& \leq p_{1} * \log_{2}{(p_{1}^{-2}*\sum_{j=1}^{n}{p_{1, j}}^{2}}) + \ldots + p_{m} * \log_{2}{(p_{m}^{-2}*\sum_{j=1}^{n}{p_{m, j}^{2}})} \\
& \leq p_{1} * \log_{2}{(p_{1}^{-1}*\max_{j}{p_{1, j}})} + \ldots + p_{m} * \log_{2}{(p_{m}^{-1}*\max_{j}{p_{m, j}})}
\end{align}
This is because $\forall i = 1, \ldots, m$:
\begin{align}
\sum_{j=1}^{n}{p_{i, j}^{2}} & = p_{i, 1} * p_{i, 1} + \ldots + p_{i, n} * p_{i, n} \\
& \leq p_{i, 1} * \max_{j}{p_{i, j}} + \ldots + p_{i, n} * \max_{j}{p_{i, j}} \\
& \leq (p_{i, 1} + \ldots + p_{i, n}) * \max_{j}{p_{i, j}} \\
& = p_{i} * \max_{j}{p_{i, j}}
\end{align}
To continue, we have:
\begin{align}
\Rightarrow 2^{-H(Y|X)} & \leq 2^{p_{1} * \log_{2}{(p_{1}^{-1}*\max_{j}{p_{1, j}})} + \ldots + p_{m} * \log_{2}{(p_{m}^{-1}*\max_{j}{p_{m, j}})}} \\
& \leq (p_{1}^{-1}*\max_{j}{p_{1, j}})^{p_{1}} * \ldots * (p_{m}^{-1}*\max_{j}{p_{m, j}})^{p_{m}} \\
& \leq \max_{j}{p_{1, j}} + \ldots + \max_{j}{p_{m, j}} ~~~~// \ \because \text{Weighted AM-GM Inequality} \\
& = \text{accuracy}(Y|X)
\end{align}
which completes the proof.
\end{proof}

\hide{
\begin{proof}
Let $Y$ and $X$ be two discrete random variables with $n$ outcomes ($y_1, \ldots, y_n$) and $m$ outcomes ($x_1, \ldots, x_m$), respectively, then their joint probabilities are ($p_{1, 1}, \ldots, p_{m, n}$), where:
\begin{equation}
    1 = p_{1} + \ldots + p_{m} = \sum_{j=1}^{n}{p_{1, j}} + \ldots + \sum_{j=1}^{n}{p_{m, j}}
\end{equation}
The accuracy is:
\begin{equation}
    \text{accuracy}(Y | X) = \sum_{i=1}^{m}{p_{i} * \max_{j}{p_{i, j}}}
\end{equation}
The conditional entropy $H(Y | X)$ is:
\begin{align}
H(Y | X) & = p_{1} * (-\sum_{j=1}^{n}{p_{1, j} * \log_{2}{p_{1, j}}}) + \ldots + p_{m} * (-\sum_{j=1}^{n}{p_{m, j} * \log_{2}{p_{m, j}}})
\end{align}
Thus we have:
\begin{align}
\Rightarrow -H(Y | X) & = p_{1} * (\sum_{j=1}^{n}{p_{1, j} * \log_{2}{p_{1, j}}}) + \ldots + p_{m} * (\sum_{j=1}^{n}{p_{m, j} * \log_{2}{p_{m, j}}}) \\
& \leq p_{1} * \log_{2}{(\sum_{j=1}^{n}{p_{1, j}^{2}})} + \ldots + p_{m} * \log_{2}{(\sum_{j=1}^{n}{p_{m, j}^{2}})} ~~~~// \ \because \text{Jensen's Inequality} \\
& \leq p_{1} * \log_{2}{(p_{1} \max_{j}{p_{1, j}})} + \ldots + p_{m} * \log_{2}{(p_{m} \max_{j}{p_{m, j}})} \\
\Rightarrow 2^{-H(Y|X)} & \leq 2^{p_{1} * \log_{2}{(p_{1} \max_{j}{p_{1, j}})} + \ldots + p_{m} * \log_{2}{(p_{m} \max_{j}{p_{m, j}})}} \\
& \leq (p_{1} \max_{j}{p_{1, j}})^{p_{1}} * \ldots * (\max_{j}{p_{m} p_{m, j}})^{p_{m}} \\
& \leq p_{1}^{2} * \max_{j}{p_{1, j}} + \ldots + p_{m}^{2} * \max_{j}{p_{m, j}} ~~~~// \ \because \text{Weighted AM-GM Inequality} \\
& \leq p_{1} * \max_{j}{p_{1, j}} + \ldots + p_{m} * \max_{j}{p_{m, j}} \\
& = \text{accuracy}(Y|X)
\end{align}
which completes the proof.
\end{proof}
}

% https://towardsdatascience.com/marginal-joint-and-conditional-probabilities-explained-by-data-scientist-4225b28907a4
% https://math.stackexchange.com/questions/1148512/how-to-get-this-upper-bound-of-this-sum-of-squares
% https://artofproblemsolving.com/wiki/index.php/AM-GM_Inequality#Power_Mean_Inequality

\subsection{Proof of Lemma~\ref{lem:cheapcm}}
\label{app:proof3}
\begin{proof}
The goal is to maximize the similarity of nodes connected by edges.
If we start from cosine similarity and L$2$-normalize the node embeddings $\mathbf{z}$, we have:
\begin{equation}
\begin{split}
& \frac{\mathbf{z}_{i}\boldsymbol{H}\mathbf{z}_{j}^{\intercal}}{\| \mathbf{z}_{i} \| \| \mathbf{z}_{j} \|} = 1, \forall (i, j) \in \mathcal{E} \\
\Rightarrow & \mathbf{z}_{i}\boldsymbol{H}\mathbf{z}_{j}^{\intercal} = \| \mathbf{z}_{i} \| \| \mathbf{z}_{j} \| = 1 \\
\Rightarrow & \mathbf{z}_{i}\boldsymbol{H} = \mathbf{z}_{j}
\end{split}
\end{equation}
Setting $\mathbf{z}_{i}$ as the input data and $\mathbf{z}_{j}$ as the target data, this equation can be solved by $d$-target linear regression with $d$ coefficients, which has the closed-form solution.
\end{proof}

\subsection{Proof of Lemma~\ref{lem:expcm}}
\label{app:proof4}
\begin{proof}

First, we rewrite the adjusted node similarity $s$ from matrix form into a simple computation:
\begin{equation}
\begin{split}
s(\mathbf{z}_{i}, \mathbf{z}_{j}, \boldsymbol{H}) & = \mathbf{z}_{i}\boldsymbol{H}\mathbf{z}_{j}^{\intercal} \\
 & =
\begin{bmatrix} \mathbf{z}_{i,1} & \cdots & \mathbf{z}_{i,d} \end{bmatrix}
\begin{bmatrix} \boldsymbol{H}_{1,1} & \cdots & \boldsymbol{H}_{1,d} \\ \vdots & \ddots & \vdots \\ \boldsymbol{H}_{d,1} & \cdots & \boldsymbol{H}_{d,d} \end{bmatrix}
\begin{bmatrix} \mathbf{z}_{j,1} \\ \vdots \\ \mathbf{z}_{j,d} \end{bmatrix} \\
 & =
\begin{bmatrix} \mathbf{z}_{i,1}\mathbf{z}_{j,1} & \cdots & \mathbf{z}_{i,1}\mathbf{z}_{j,d} \\ \vdots & \ddots & \vdots \\ \mathbf{z}_{i,d}\mathbf{z}_{j,1} & \cdots & \mathbf{z}_{i,d}\mathbf{z}_{j,d} \end{bmatrix}
\odot
\begin{bmatrix} \boldsymbol{H}_{1,1} & \cdots & \boldsymbol{H}_{1,d} \\ \vdots & \ddots & \vdots \\ \boldsymbol{H}_{d,1} & \cdots & \boldsymbol{H}_{d,d} \end{bmatrix} \\
 & =
\mathbf{z}_{i,1}\mathbf{z}_{j,1}\boldsymbol{H}_{1,1} + 
\mathbf{z}_{i,1}\mathbf{z}_{j,2}\boldsymbol{H}_{1,2} + \cdots +
\mathbf{z}_{i,1}\mathbf{z}_{j,d}\boldsymbol{H}_{1,d} + \cdots +
\mathbf{z}_{i,d}\mathbf{z}_{j,d}\boldsymbol{H}_{d,d}
\end{split}
\end{equation}

Next, to maximize the similarity of nodes connected by positive edges, and to minimize the similarity of nodes connected by negative edges, we have:
\begin{equation}
s(\mathbf{z}_{i}, \mathbf{z}_{j}, \boldsymbol{H}) = 
\begin{cases}
1 & (i, j) \in \mathcal{E} \\
0 & (i, j) \in \mathcal{E}_{\text{neg}}
\end{cases}
\end{equation}
Therefore, this equation can be solved by linear regression with $d^{2}$ coefficients, which has the closed-form solution.

\end{proof}

\subsection{Proof of Lemma~\ref{lem:time}}
\label{app:proof5}
\begin{proof}
\method includes four parts: SVD, PCA, compatibility matrix estimation, and LogitReg.
The time complexity of truncated SVD is $O(d^{2}|\mathcal{V}|)$, and the one of PCA is $f^{2}|\mathcal{V}|+f^{3}$.
Compatibility matrix estimation is optimized by Ridge regression with regularized least-squares routine, whose time complexity is $d^{4}|\mathcal{E}|$.
The time complexity of training LogitReg is $dt|\mathcal{E}|$, where $t$ is the number of epochs.
In our experiments, $t$ is no greater than $100$, and $|\mathcal{V}|$ is much less than $|\mathcal{E}|$ in most datasets.
By combining all the terms and keeping only the dominant ones, we have time complexity of \method for link prediction $O(f^{2}|\mathcal{V}| + f^{3} + d^{4}|\mathcal{E}|)$.
\end{proof}

\section{Appendix: Algorithms}

\begin{algorithm}[h]
    \caption{Compatibility Matrix with Negative Edges \label{algo:a1}}
    \KwIn{Preprocessed node embedding $\hat{\mathbf{Z}}$, edge set $\mathcal{E}$, and sample size $S$}
    Extract $2$-core edge set $\mathcal{E}_{pos}$ from $\mathcal{E}$\;
    Estimate $\mathbf{H}$ with $\hat{\mathbf{Z}}$ and $\mathcal{E}_{pos}$ by Lemma~\ref{lem:cheapcm}\;
    \If{$|\mathcal{E}_{pos}| > S$}{
        Sample $S$ edges from $\mathcal{E}_{pos}$\;
    }
    Initialize $\mathbf{H}^{*}$ with $\mathbf{H}$\;
    Keep top coefficients in upper triangle of $\mathbf{H}^{*}$ with $95\%$ energy\;
    Sample $2|\mathcal{E}_{pos}|$ negative edges as $\mathcal{E}_{neg}$\;
    Estimate $\mathbf{H}^{*}$ with $\hat{\mathbf{Z}}$, $\mathcal{E}_{pos}$ and $\mathcal{E}_{neg}$ by Lemma~\ref{lem:expcm}\;
    Return $\mathbf{H}^{*}$\;
\end{algorithm}

\begin{algorithm}[h]
    \caption{\analysis for Link Prediction \label{algo:a2}}
    \KwIn{Node embedding $\mathbf{Z}$, train edge set $\mathcal{E}_{train}$, valid edge set $\mathcal{E}_{valid}$, valid edge labels $\mathbf{y}_{valid}$, sample size $S$, and bin size $k$}
    Preprocess $\mathbf{Z}$ into $\hat{\mathbf{Z}}$ by column-wise standardization and row-wise L$2$-normalization\;
    $\mathbf{H}^{*} = \text{Compatibility-Matrix-with-Negative-Edges}(\hat{\mathbf{Z}}, \mathcal{E}_{train}, S)$\;
    \If{$|\mathcal{E}_{pos}| > S$}{
        Sample $S$ edges from $\mathcal{E}_{pos}$\;
    }
    \Else{
        $\mathcal{E}_{pos} \leftarrow \mathcal{E}$\;
    }
    Sample $2|\mathcal{E}_{pos}|$ negative edges as $\mathcal{E}_{neg}$\;
    Fit $k$-bins discretizer with $\hat{\mathbf{z}}_{i}\mathbf{H}^{*}\hat{\mathbf{z}}_{j}^{\intercal}, \forall (i, j) \in \mathcal{E}_{pos} \cup \mathcal{E}_{neg}$\;
    Discretize $\hat{\mathbf{z}}_{i}\mathbf{H}^{*}\hat{\mathbf{z}}_{j}^{\intercal}, \forall (i, j) \in \mathcal{E}_{valid}$ into $k$ bins as $\mathbf{s}_{valid}$\;
    Return \score, computed with $\mathbf{s}_{valid}$ and $\mathbf{y}_{valid}$ by Eq.~\ref{eqn:score}\;
\end{algorithm}

\begin{algorithm}[h]
    \caption{\analysis for Node Classification \label{algo:a3}}
    \KwIn{Train, valid, and test node embedding $\mathbf{Z}_{train}$, $\mathbf{Z}_{valid}$, and $\mathbf{Z}_{test}$, train and valid node labels $\mathbf{y}_{train}$ and $\mathbf{y}_{valid}$, and cluster number $k$}
    Preprocess $\mathbf{Z}$ into $\hat{\mathbf{Z}}$ row-wise L$2$-normalization\;
    Fit clustering model with $\hat{\mathbf{Z}}_{test}$\;
    Assign cluster labels $\mathbf{s}_{train}$ and $\mathbf{s}_{valid}$ to $\hat{\mathbf{Z}}_{train}$ and $\hat{\mathbf{Z}}_{valid}$, respectively\;
    Return \score, computed with $\mathbf{s}_{train} \cup \mathbf{s}_{valid}$ and $\mathbf{y}_{train} \cup \mathbf{y}_{valid}$ by Eq.~\ref{eqn:score}\;
\end{algorithm}

% \begin{algorithm}[t]
% \begin{minipage}[t] {0.49\linewidth}
%     \caption{\label{algo:a1}}
%     Input:
% \end{minipage} \hfill
% \begin{minipage}[t] {0.49\linewidth}
%     \caption{\label{algo:a2}}
%     Input:
% \end{minipage}
% \end{algorithm}

\section{Appendix: Experiments} \label{app:expr}

\subsection{Link Prediction in Synthetic Datasets} \label{app:lp_sc}
To study the information of the derived node embeddings, we conduct \analysis on each component, and LogitReg to have test performance on link prediction.
In Table~\ref{table:exprlp_syn_info}, the components with the top-$2$ \score as well have the top-$2$ test performance in every scenario.

\begin{table*}[h]
\caption{
    Results of each derived node embeddings. \score and test Hits@$100$ on link prediction are reported. \textcolor{red}{Red} highlights the results close to random guessing.
}
\centering{\resizebox{0.8\textwidth}{!}{
\begin{tabular}{ rc | cc | cc | cc  }
	\toprule
	\multirow{2}{*}{\textbf{Metric}} & \textbf{Feature}
	    & \textbf{Random $\mathbf{X}$}
	    & \textbf{Random $\mathbf{X}$} 
	    & \textbf{Global $\mathbf{X}$} 
        & \textbf{Global $\mathbf{X}$}
	    & \textbf{Local $\mathbf{X}$}
	    & \textbf{Local $\mathbf{X}$} \\
    & \textbf{Component}
	    & \textbf{Diagonal $\mathbf{A}$}
        & \textbf{Off-Diag. $\mathbf{A}$}
	    & \textbf{Diagonal $\mathbf{A}$} 
	    & \textbf{Off-Diag. $\mathbf{A}$} 
	    & \textbf{Diagonal $\mathbf{A}$}
	    & \textbf{Off-Diag. $\mathbf{A}$} \\
	\midrule
    \multirow{5}{*}{\score} & $\boldsymbol{C1: U}$
	    & 75.1$\pm$0.9
	    & 74.3$\pm$0.5
	    & 75.3$\pm$0.6
        & 74.1$\pm$0.7
	    & 75.2$\pm$0.8
	    & 74.8$\pm$0.5 \\
    & $\boldsymbol{C2: R}$
	    & \textbf{76.5$\pm$0.9}
	    & \textbf{76.3$\pm$0.5}
	    & 76.7$\pm$0.8
        & 76.3$\pm$0.2
	    & 76.6$\pm$0.8
	    & 76.3$\pm$0.4 \\
    & $\boldsymbol{C3: F}$
	    & \textcolor{red}{50.0$\pm$0.0}
	    & \textcolor{red}{50.0$\pm$0.0}
	    & 67.0$\pm$0.5
        & 71.5$\pm$0.6
	    & 75.9$\pm$1.1
	    & 75.5$\pm$0.6 \\
	& $\boldsymbol{C4: P}$
	    & 75.1$\pm$0.7
	    & 73.2$\pm$0.6
	    & 78.2$\pm$0.7
        & 78.5$\pm$0.5
	    & 78.5$\pm$1.2
	    & 78.9$\pm$0.5 \\
    & $\boldsymbol{C5: S}$
	    & 74.3$\pm$1.1
	    & 72.0$\pm$0.5
	    & \textbf{78.7$\pm$1.0}
        & \textbf{79.2$\pm$0.5}
	    & \textbf{79.0$\pm$0.9}
	    & \textbf{81.1$\pm$0.6} \\
    \midrule
    \multirow{5}{*}{Test Hits@$100$} & $\boldsymbol{C1: U}$
	    & \textbf{83.3$\pm$1.2}
	    & \textbf{76.4$\pm$2.1}
	    & 83.5$\pm$1.2
        & 76.7$\pm$0.5
	    & 83.2$\pm$1.1
	    & 78.2$\pm$0.9 \\
    & $\boldsymbol{C2: R}$
	    & 83.0$\pm$1.5
	    & 74.7$\pm$1.4
	    & 83.0$\pm$1.3
        & 75.0$\pm$1.5
	    & 83.2$\pm$1.0
	    & 75.3$\pm$0.7 \\
    & $\boldsymbol{C3: F}$
	    & \textcolor{red}{02.2$\pm$0.2}
	    & \textcolor{red}{02.5$\pm$0.2}
	    & 52.3$\pm$1.9
        & 63.0$\pm$2.2
	    & 80.7$\pm$1.1
	    & 77.1$\pm$1.0 \\
	& $\boldsymbol{C4: P}$
	    & 82.3$\pm$1.2
	    & 73.4$\pm$0.4
	    & 86.2$\pm$1.1
        & \textbf{79.3$\pm$0.7}
	    & 85.7$\pm$1.5
	    & 79.4$\pm$1.3 \\
    & $\boldsymbol{C5: S}$
	    & 80.7$\pm$0.9
	    & 68.1$\pm$1.8
	    & \textbf{86.4$\pm$0.9}
        & 79.2$\pm$1.1
	    & \textbf{86.5$\pm$1.0}
	    & \textbf{85.2$\pm$1.1} \\
 %     \midrule
 %    \multirow{5}{*}{Test AUC-ROC} & $\boldsymbol{U}$
	%     & 94.7$\pm$0.1
	%     & \textbf{95.7$\pm$0.3}
	%     & 94.7$\pm$0.1
 %        & 95.6$\pm$0.2
	%     & 94.7$\pm$0.1
	%     & 95.5$\pm$0.1 \\
 %    & $\boldsymbol{R}$
	%     & \textbf{94.8$\pm$0.2}
	%     & 95.3$\pm$0.2
	%     & 94.8$\pm$0.2
 %        & 95.4$\pm$0.1
	%     & 94.8$\pm$0.1
	%     & 95.3$\pm$0.2 \\
 %    & $\boldsymbol{F}$
	%     & \textcolor{red}{50.3$\pm$0.5}
	%     & \textcolor{red}{49.8$\pm$0.6}
	%     & 90.0$\pm$0.3
 %        & 93.0$\pm$0.3
	%     & 94.6$\pm$0.1
	%     & 95.0$\pm$0.4 \\
	% & $\boldsymbol{P}$
	%     & 94.4$\pm$0.1
	%     & 95.0$\pm$0.2
	%     & 96.2$\pm$0.2
 %        & 96.1$\pm$0.1
	%     & 95.7$\pm$0.1
	%     & 96.3$\pm$0.2 \\
 %    & $\boldsymbol{S}$
	%     & 93.9$\pm$0.3
	%     & 93.6$\pm$0.1
	%     & \textbf{96.5$\pm$0.1}
 %        & \textbf{96.7$\pm$0.1}
	%     & \textbf{96.0$\pm$0.2}
	%     & \textbf{97.1$\pm$0.2} \\
	\bottomrule
\end{tabular}
}}
\label{table:exprlp_syn_info}
\end{table*}

\subsection{Link Prediction in Real-World Datasets}
To study the effectiveness of each derived node embedding, we conduct experiments on each individual component for the real-world datasets.
As shown in Table~\ref{tab:exprlp_comp}, different components have variable impact depending on the input graph.

\begin{table*}[h]
\caption{
    \emphasize{Ablation study} - \method on each component of derived node embedding. 
    Hits@$100$ is reported for most datasets, and Hits@$1000$ for the large datasets (Products, Twitch, and Pokec).
}
\centering{\resizebox{\textwidth}{!}{
\begin{tabular}{ l | ccccccc | ccccc }
	\toprule
	\textbf{Model}
	    & \textbf{Cora}
	    & \textbf{CiteSeer} 
	    & \textbf{PubMed} 
	    & \textbf{Comp.}
	    & \textbf{Photo}
	    & \textbf{ArXiv}
	    & \textbf{Products}
	    & \textbf{Cham.} 
	    & \textbf{Squirrel} 
	    & \textbf{Actor} 
	    & \textbf{Twitch}
	    & \textbf{Pokec} \\
	\midrule
	$C1: U$
	    & 48.3$\pm$0.8
	    & 36.9$\pm$2.5
	    & 43.7$\pm$1.8
	    & 24.0$\pm$1.8
	    & 38.5$\pm$2.6
	    & 18.1$\pm$0.6
	    & 13.2$\pm$0.3
	    & 75.0$\pm$5.2
	    & 12.3$\pm$1.8
	    & 23.2$\pm$2.2
	    & \textbf{15.8$\pm$0.2}
	    & \textbf{16.2$\pm$0.6} \\
	$C2: R$
	    & 61.5$\pm$1.2
	    & 50.0$\pm$1.9
	    & 47.9$\pm$0.8
	    & 19.4$\pm$0.8
	    & 36.8$\pm$3.3
	    & 12.3$\pm$1.0
	    & 09.4$\pm$0.7
	    & 64.6$\pm$3.2
	    & 08.2$\pm$1.4
	    & \textbf{34.2$\pm$2.3}
	    & 01.5$\pm$0.2
	    & 05.2$\pm$0.2 \\
    $C3: F$
	    & 58.3$\pm$3.2
	    & 71.5$\pm$2.4
	    & 41.6$\pm$0.4
	    & 07.1$\pm$0.4
	    & 15.6$\pm$1.2
	    & 04.9$\pm$0.2
	    & 00.4$\pm$0.1
	    & 10.7$\pm$1.0
	    & 00.6$\pm$0.1
	    & 02.6$\pm$0.5
	    & 01.7$\pm$0.6
	    & 00.5$\pm$0.2 \\
	$C4: P$
	    & 67.3$\pm$1.8
	    & 60.9$\pm$2.5
	    & 48.1$\pm$1.8
	    & \textbf{28.6$\pm$2.4}
	    & \textbf{42.4$\pm$1.4}
	    & \textbf{34.2$\pm$1.0}
	    & 27.6$\pm$0.5
	    & \textbf{84.3$\pm$1.8}
	    & \textbf{20.9$\pm$2.3}
	    & 13.1$\pm$1.1
	    & 07.3$\pm$1.0
	    & 12.4$\pm$0.8 \\
    $C5: S$
	    & \textbf{82.4$\pm$1.0}
	    & \textbf{88.7$\pm$1.5}
	    & \textbf{63.3$\pm$1.1}
	    & 29.0$\pm$2.3
	    & 40.3$\pm$1.1
	    & 33.6$\pm$1.1
	    & \textbf{28.1$\pm$0.7}
	    & 80.5$\pm$2.7
	    & 19.2$\pm$1.2
	    & 22.3$\pm$0.9
	    & 02.7$\pm$1.8
	    & \textbf{16.2$\pm$0.6} \\
	\bottomrule
\end{tabular}
}}
\label{tab:exprlp_comp}
\end{table*}

\subsection{Sensitivity Analysis of \analysis in Link Prediction}
\begin{figure}[h]
\begin{minipage}{0.62\textwidth}
\captionof{table}{
    \emphasize{Sensitivity analysis} of \analysis in link prediction. \score is reported.
}
\centering{\resizebox{1\textwidth}{!}{
\begin{tabular}{ c | ccccc  }
	\toprule
	\textbf{Number of Bins}
	    & $\mathbf{k=4}$
	    & $\mathbf{k=8}$
        & $\mathbf{k=16}$
	    & $\mathbf{k=32}$
	    & $\mathbf{k=64}$ \\
	\midrule
    $C1: U$
	    & 74.2$\pm$0.3
	    & 79.7$\pm$0.2
	    & 80.4$\pm$0.3
	    & 80.8$\pm$0.3
	    & 81.0$\pm$0.3 \\
    $C2: R$
	    & 76.5$\pm$0.2
	    & 80.4$\pm$0.0
	    & 81.2$\pm$0.3
	    & 81.4$\pm$0.3
	    & 81.5$\pm$0.3 \\
    $C3: F$
	    & 63.8$\pm$0.1
	    & 64.9$\pm$0.2
	    & 65.0$\pm$0.1
	    & 65.0$\pm$0.1
	    & 65.1$\pm$0.1 \\
	$C4: P$
	    & 77.3$\pm$0.2
	    & 82.1$\pm$0.1
	    & 83.0$\pm$0.3
	    & 83.3$\pm$0.3
	    & 83.4$\pm$0.3 \\
    $C5: S$
	    & 77.3$\pm$0.1
	    & 82.5$\pm$0.1
	    & 83.3$\pm$0.1
	    & 83.6$\pm$0.2
	    & 83.7$\pm$0.2 \\
	\bottomrule
\end{tabular}
}}
\label{tab:sensitivity}
\end{minipage}
\hfill
\begin{minipage}{0.35\textwidth}
\centering
\includegraphics[scale=0.5]{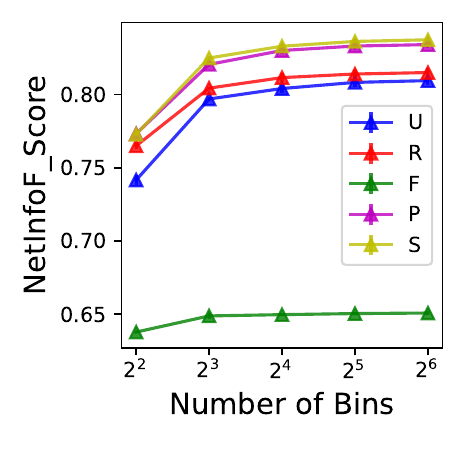}
\vspace{-2mm}
\captionof{figure}{\emphasize{Sensitivity analysis} - \analysis is insensitive to the number of bins $k$.}
\label{fig:sensitivity}
\end{minipage}
\end{figure}

We conduct a sensitivity analysis on a medium size dataset ``Computers'', to study the effect of the number of bins $k$ in \analysis (Algo.~\ref{algo:a2}).
As shown in Table~\ref{tab:sensitivity} and Fig.~\ref{fig:sensitivity}, \score is insensitive to the exact value of $k$, forming a plateau when $k$ increases.

\subsection{Comparison with Subgraph GNN in Real-World Datasets}
Although the comparison with subgraph GNN is beyond the scope of this paper, the experimental results of a subgraph GNN, SEAL \citep{zhang2018link}, are provided in Table~\ref{tab:exprlp_seal} for the interested readers.
The hidden size is set to $128$ as the same as other baselines, and the hyperparameter search is done on the number of hops ($[1, 3]$), learning rate ($[0.001, 0.01]$), weight decay ($[0, 5e^{-4}]$), and the number of layers ($[2, 3]$).
Other hyperparameters are set as the default ones in \citet{zhang2021labeling}.

To emphasize again, as shown in Table~\ref{tab:salesman}, subgraph GNNs are beyond the scope of this paper, because:
(1) they are not designed as general GNNs and can only solve link prediction;
(2) they are only scalable to small graphs.
To give a concrete example, using an AWS EC2 G4dn instance with $384$GB RAM and running on the largest dataset Photo without running out of memory (O.O.M.) in the preprocessing step, SEAL takes $17606$ seconds (around $4.9$ hours) to train on a single split of data, while \method takes only $52$ seconds, which is $\mathbf{339\times}$ faster than SEAL.

\begin{table*}[h]
\caption{
    Comparison with SEAL. Hits@$100$ is reported for most datasets, and Hits@$1000$ for the large datasets (Products, Twitch, and Pokec).
}
\centering{\resizebox{\textwidth}{!}{
\begin{tabular}{ l | ccccccc | ccccc }
	\toprule
	\textbf{Model}
	    & \textbf{Cora}
	    & \textbf{CiteSeer} 
	    & \textbf{PubMed} 
	    & \textbf{Comp.}
	    & \textbf{Photo}
	    & \textbf{ArXiv}
	    & \textbf{Products}
	    & \textbf{Cham.} 
	    & \textbf{Squirrel} 
	    & \textbf{Actor} 
	    & \textbf{Twitch}
	    & \textbf{Pokec} \\
	\midrule
	SEAL
	    & 66.4$\pm$0.6
	    & 61.4$\pm$1.0
	    & \textbf{62.8$\pm$1.7}
	    & \textcolor{red}{O.O.M.}
	    & \textbf{55.5$\pm$1.0}
	    & \textcolor{red}{O.O.M.}
	    & \textcolor{red}{O.O.M.}
	    & \textbf{89.4$\pm$1.0}
	    & \textcolor{red}{O.O.M.}
	    & \textbf{43.6$\pm$0.7}
	    & \textcolor{red}{O.O.M.}
	    & \textcolor{red}{O.O.M.} \\
	\method 
        & \textbf{81.3$\pm$0.6}
	    & \textbf{87.3$\pm$1.3}
	    & 59.7$\pm$1.1
	    & \textbf{31.1$\pm$1.9}
	    & 46.8$\pm$2.2
	    & \textbf{39.2$\pm$1.8}
	    & \textbf{35.2$\pm$1.1}
	    & 86.9$\pm$2.3
	    & \textbf{24.2$\pm$2.0}
	    & 36.2$\pm$1.2
	    & \textbf{19.6$\pm$0.7}
	    & \textbf{31.3$\pm$0.5} \\
	\bottomrule
\end{tabular}
}}
\label{tab:exprlp_seal}
\end{table*}

\subsection{Node Classification in Synthetic Datasets} \label{app:nc_syn}
We keep the same settings of sanity checks in \textsc{SlimG}, but use our graph generation process in Appx.~\ref{app:data}.
To study the information of the derived node embeddings, we conduct \analysis on each component, and LogitReg to have test performance on node classification.
In Table~\ref{table:exprnc_syn_info}, the component with the highest \score as well has the highest test accuracy in every scenario.
In Table~\ref{table:exprnc_syn}, \method generalizes to all scenarios as \textsc{SlimG} does.

\begin{table*}[h]
\caption{
    \emphasize{\method works well} on node classification in synthetic datasets. 
    Accuracy is reported.
}
\centering{\resizebox{0.7\textwidth}{!}{
\begin{tabular}{ l | c | cc | cc }
	\toprule
	\multirow{2}{*}{\textbf{Model}}
	    & \textbf{Useful $\mathbf{X}$}
	    & \textbf{Random $\mathbf{X}$} 
	    & \textbf{Random $\mathbf{X}$} 
	    & \textbf{Useful $\mathbf{X}$}
	    & \textbf{Useful $\mathbf{X}$} \\
	    & \textbf{Uniform $\mathbf{A}$}
	    & \textbf{Homophily $\mathbf{A}$} 
	    & \textbf{Heterophily $\mathbf{A}$} 
	    & \textbf{Homophily $\mathbf{A}$}
	    & \textbf{Heterophily $\mathbf{A}$} \\
	\midrule
    \textsc{SlimG}
	    & 85.4$\pm$2.3
        & 88.9$\pm$0.4
        & 87.4$\pm$2.5
        & 97.3$\pm$0.2
        & 97.0$\pm$0.1 \\
    \midrule
	\method
	    & 86.7$\pm$1.5
	    & 88.6$\pm$1.5
	    & 87.9$\pm$1.5
	    & 97.1$\pm$0.1
	    & 97.0$\pm$0.2 \\
	\bottomrule
\end{tabular}
}}
\label{table:exprnc_syn}
\end{table*}

\begin{table*}[h]
\caption{
    Results of each derived node embeddings. \score, and test accuracy on node classification are reported. \textcolor{red}{Red} highlights the results close to random guessing.
}
\centering{\resizebox{0.8\textwidth}{!}{
\begin{tabular}{ rc | c | cc | cc  }
	\toprule
	\multirow{2}{*}{\textbf{Metric}} & \textbf{Feature}
	    & \textbf{Useful $\mathbf{X}$}
	    & \textbf{Random $\mathbf{X}$} 
        & \textbf{Random $\mathbf{X}$}
	    & \textbf{Useful $\mathbf{X}$}
	    & \textbf{Useful $\mathbf{X}$} \\
    & \textbf{Component}
        & \textbf{Uniform $\mathbf{A}$}
	    & \textbf{Homophily $\mathbf{A}$} 
	    & \textbf{Heterophily $\mathbf{A}$} 
	    & \textbf{Homophily $\mathbf{A}$}
	    & \textbf{Heterophily $\mathbf{A}$} \\
	\midrule
    \multirow{5}{*}{\score} & $\boldsymbol{C1: U}$
	    & \textcolor{red}{25.9$\pm$0.2}
	    & \textbf{73.8$\pm$2.8}
	    & \textbf{74.4$\pm$3.9}
	    & 73.6$\pm$2.6
	    & 75.4$\pm$4.5 \\
    & $\boldsymbol{C2: R}$
	    & \textcolor{red}{25.7$\pm$0.1}
	    & 62.5$\pm$3.5
	    & 56.7$\pm$1.3
	    & 64.7$\pm$3.7
	    & 56.3$\pm$3.3 \\
    & $\boldsymbol{C3: F}$
	    & \textbf{64.6$\pm$4.2}
	    & \textcolor{red}{26.0$\pm$0.4}
	    & \textcolor{red}{26.0$\pm$0.4}
	    & 73.1$\pm$3.2
	    & 73.1$\pm$3.2 \\
	& $\boldsymbol{C4: P}$
	    & 33.7$\pm$1.5
	    & 56.0$\pm$4.2
	    & 58.9$\pm$3.9
	    & \textbf{87.2$\pm$3.6}
	    & 87.2$\pm$3.2 \\
    & $\boldsymbol{C5: S}$
	    & 33.9$\pm$0.8
	    & 58.7$\pm$4.8
	    & 55.5$\pm$3.5
	    & 86.1$\pm$2.4
	    & \textbf{88.3$\pm$3.4} \\    
    \midrule
    \multirow{5}{*}{Test Accuracy} & $\boldsymbol{C1: U}$
	    & \textcolor{red}{25.1$\pm$0.3}
	    & \textbf{80.3$\pm$0.9}
	    & \textbf{81.3$\pm$0.8}
	    & 80.3$\pm$0.9
	    & 81.3$\pm$0.8 \\
    & $\boldsymbol{C2: R}$
	    & \textcolor{red}{25.3$\pm$0.4}
	    & 60.7$\pm$0.8
	    & 58.4$\pm$1.4
	    & 61.5$\pm$0.9
	    & 58.8$\pm$1.3 \\
    & $\boldsymbol{C3: F}$
	    & \textbf{74.7$\pm$0.8}
	    & \textcolor{red}{25.4$\pm$0.6}
	    & \textcolor{red}{25.4$\pm$0.6}
	    & 74.7$\pm$1.1
	    & 74.7$\pm$1.1 \\
	& $\boldsymbol{C4: P}$
	    & 47.9$\pm$0.9
	    & 75.2$\pm$0.9
	    & 77.4$\pm$0.8
	    & \textbf{97.1$\pm$0.1}
	    & 96.6$\pm$0.1 \\
    & $\boldsymbol{C5: S}$
	    & 47.9$\pm$0.9
	    & 75.8$\pm$1.0
	    & 74.1$\pm$0.8
	    & 96.9$\pm$0.1
	    & \textbf{96.9$\pm$0.1} \\
	\bottomrule
\end{tabular}
}}
\label{table:exprnc_syn_info}
\end{table*}

\subsection{Node Classification in Real-World Datasets} \label{app:nc_real}
We follow the same experimental settings in \textsc{SlimG}.
The nodes are randomly split by the ratio $2.5\%/2.5\%/95\%$ into the training, validation and testing sets.
% Since our contribution majorly focuses on improving the performance on link prediction, to save space, we only compare \method with the top baseline \textsc{SlimG} on node classification.
In Table~\ref{table:exprnc}, we find that \method always wins and ties with the top baselines.

\begin{table*}[h]
\caption{
    \emphasize{\method works well} on node classification in real-world datasets. 
    Accuracy is reported.
}
\centering{\resizebox{1\textwidth}{!}{
\begin{tabular}{ l | ccccccc | ccccc }
	\toprule
	\textbf{Model}
	    & \textbf{Cora}
	    & \textbf{CiteSeer} 
	    & \textbf{PubMed} 
	    & \textbf{Comp.}
	    & \textbf{Photo}
	    & \textbf{ArXiv}
	    & \textbf{Products}
	    & \textbf{Cham.} 
	    & \textbf{Squirrel} 
	    & \textbf{Actor} 
	    & \textbf{Twitch}
	    & \textbf{Pokec} \\
	\midrule
    GCN
	  & 76.0$\pm$1.2
	    & 65.0$\pm$2.9
	    & 84.3$\pm$0.5
	    & 85.1$\pm$0.9
	    & 91.6$\pm$0.5
	    & 62.8$\pm$0.6
	    & O.O.M.
	    & 38.5$\pm$3.0
	    & \textbf{31.4$\pm$1.8}
	    & 26.8$\pm$0.4
	    & 57.0$\pm$0.1
	    & 63.9$\pm$0.4 \\
	SAGE
	    & 74.6$\pm$1.3
	    & 63.7$\pm$3.6
	    & 82.9$\pm$0.4
	    & 83.8$\pm$0.5
	    & 90.6$\pm$0.5
	    & 61.5$\pm$0.6
	    & O.O.M.
	    & 39.8$\pm$4.3
	    & 27.0$\pm$1.3
	    & 27.8$\pm$0.9
	    & 56.6$\pm$0.4
	    & 68.9$\pm$0.1 \\
    H$^{2}$GCN
	    & 77.6$\pm$0.9
	    & 64.7$\pm$3.8
	    & \textbf{85.4$\pm$0.4}
	    & 49.5$\pm$16.
	    & 75.8$\pm$11.
	    & O.O.M.
	    & O.O.M.
	    & 31.9$\pm$2.6
	    & 25.0$\pm$0.5
	    & 28.9$\pm$0.6
	    & 58.7$\pm$0.0
	    & O.O.M. \\
    GPR-GNN
	    & \textbf{78.8$\pm$1.3}
	    & 64.2$\pm$4.0
	    & 85.1$\pm$0.7
	    & 85.0$\pm$1.0
	    & \textbf{92.6$\pm$0.3}
	    & 58.5$\pm$0.8
	    & O.O.M.
	    & 31.7$\pm$4.7
	    & 26.2$\pm$1.6
	    & 29.5$\pm$1.1
	    & 57.6$\pm$0.2
	    & 67.6$\pm$0.1 \\
	GAT
	    & 78.2$\pm$1.2
	    & 65.8$\pm$4.0
	    & 83.6$\pm$0.2
	    & 85.4$\pm$1.4
	    & 91.7$\pm$0.5
	    & 58.2$\pm$1.0
	    & O.O.M.
	    & 39.1$\pm$4.1
	    & 28.6$\pm$0.6
	    & 26.4$\pm$0.4
	    & O.O.M.
	    & O.O.M. \\
    \textsc{SlimG}
	    & 77.8$\pm$1.1
	    & \textbf{67.1$\pm$2.3}
	    & 84.6$\pm$0.5
	    & 86.3$\pm$0.7
	    & 91.8$\pm$0.5
	    & 66.3$\pm$0.3
	    & 84.9$\pm$0.0
	    & 40.8$\pm$3.2
	    & 31.1$\pm$0.7
	    & \textbf{30.9$\pm$0.6}
	    & 59.7$\pm$0.1
	    & 73.9$\pm$0.1 \\
    \midrule
	\method
	    & 77.5$\pm$1.3
	    & 64.5$\pm$3.6
	    & 84.1$\pm$0.5
	    & \textbf{86.6$\pm$0.6}
	    & 91.6$\pm$0.3
	    & \textbf{66.8$\pm$0.7}
	    & \textbf{85.3$\pm$0.0}
	    & \textbf{41.6$\pm$3.2}
	    & 30.4$\pm$1.5
	    & 30.7$\pm$0.3
	    & \textbf{61.0$\pm$0.2}
	    & \textbf{74.0$\pm$0.1} \\
	\bottomrule
\end{tabular}
}}
\label{table:exprnc}
\end{table*}

\section{Appendix: Reproducibility} \label{app:rep}

\subsection{Synthetic Datasets} \label{app:data}
The synthetic datasets are composed of two parts, namely, graph structure and node features.
Noting that link prediction and node classification share the same generator of graph structure, but differ in the one of node features.
Noises are randomly injected into both graph structure and node features to ensure that they contain consistent usable information across different scenarios.
The number of nodes is set to be $4000$, and the number of features is set to be $800$.

\textbf{Graph Structure.}
There are three kinds of graph structures, namely, diagonal, off-diagonal, and uniform.
For each graph, we equally assign labels with $c$ classes to all nodes.
In the diagonal structure, the nodes are connected to the nodes with the same class label, which exhibits homophily.
In the off-diagonal structure, the nodes are connected to the nodes with one different class label, which exhibits heterophily.
In the uniform structure, the connections are randomly made between nodes.
Other than randomly picking node pairs to make connections, we mimic the phenomenon that the nodes are connected by higher-order structure in the real-world \citep{eswaran2020higher}.
To achieve that, in the diagonal structure, a random amount of nodes (between $4$ and $8$) with the same class are randomly picked and made into a clique;
In the off-diagonal structure, a random amount of nodes (between $4$ and $8$) from each of the two classes are randomly picked and made into a bipartite clique.
This process continues until the graph reaches our desired density.
In the link prediction, the assigned node labels are not used; in the node classification, they are used as the target labels.

\textbf{Node Features in Link Prediction.}
In the case that the node features are useful, they are generated by the left singular vectors of the $2$-step random walk matrix.
The $(i, j)$ element of the matrix is the counting of the node on row $i$ visited the node on column $j$, and each node walks for $1000$ trials.
The node features are directly used in the global scenarios, but split into different slices based on the labels in the local scenarios.
The random node features are the rows in a random binary matrix.

\textbf{Node Features in Node Classification.}
In the case that the node features are useful, we randomly sample a center for each class label.
For nodes with the same class, we add Gaussian noises on top of their class center.
The random node features are the rows in a random binary matrix.

\subsection{Real-World Datasets}
In the experiments, we use $7$ homophily and $5$ heterophily real-world datasets that have been widely used before.
All the graphs are made undirected and their statistics are reported in Table~\ref{table:datastat}.
We also conduct experiments on $3$ link prediction datasets from Open Graph Benchmark (OGB) \citep{hu2020ogb}, namely ogbl-ddi\footnote{\url{https://ogb.stanford.edu/docs/linkprop/\#ogbl-ddi}}, ogbl-collab\footnote{\url{https://ogb.stanford.edu/docs/linkprop/\#ogbl-collab}}, and ogbl-ppa\footnote{\url{https://ogb.stanford.edu/docs/linkprop/\#ogbl-ppa}}.
% and ogbl-citation2\footnote{\url{https://ogb.stanford.edu/docs/linkprop/\#ogbl-citation2}}
% For the success story, we use a heterogeneous shopping queries network \citep{reddy2022shopping}\footnote{\url{https://github.com/amazon-science/esci-data}}, which contains four edge types.

\textbf{Homophily Graphs.}
Cora \citep{motl2015ctu}\footnote{\url{https://relational.fit.cvut.cz/dataset/CORA}}, 
CiteSeer \citep{rossiaaai15}\footnote{\url{https://linqs.org/datasets/\#citeseer-doc-classification}}
, and PubMed \citep{PubMed}\footnote{\url{https://www.nlm.nih.gov/databases/download/pubmed_medline.html}}
are citation networks between research articles.
Computers and Photo \citep{ni2019justifying}\footnote{\url{https://nijianmo.github.io/amazon/index.html}} are Amazon co-purchasing networks between products.
ogbn-arXiv and ogbn-Products are large graphs from OGB \citep{hu2020ogb}.
ogbn-arXiv\footnote{\url{https://ogb.stanford.edu/docs/nodeprop/\#ogbn-arxiv}} is a citation network of papers from arXiv; and ogbn-Products\footnote{\url{https://ogb.stanford.edu/docs/nodeprop/\#ogbn-products}} is also an Amazon product co-purchasing network.
In the node classification task, we omit the classes with instances fewer than $100$ so that each class has enough training data.

\textbf{Heterophily Graphs.}
Chameleon and Squirrel \citep{musae}\footnote{\url{https://github.com/benedekrozemberczki/MUSAE/}} are Wikipedia page-to-page networks between articles from Wikipedia. 
Actor \citep{pei2020geom}\footnote{\url{https://github.com/graphdml-uiuc-jlu/geom-gcn/tree/master/new_data/film}} is a co-occurrence network of actors on Wikipedia pages.
Twitch \citep{musae}\footnote{\url{https://github.com/benedekrozemberczki/datasets\#twitch-social-networks}} and Pokec \citep{takac2012data, leskovec2014snap}\footnote{\url{https://snap.stanford.edu/data/soc-Pokec.html}} are online social networks, which are relabeled by \citet{Lim21LINKX}\footnote{\url{https://github.com/CUAI/Non-Homophily-Large-Scale}} to present heterophily.
Penn94 is not included in this paper because of legal issues.

\begin{table}[h]
\caption{
    Network Statistics. The left and right parts are homophily and heterophily, respectively.
}
\centering{\resizebox{1\textwidth}{!}{
\begin{tabular}{l | rrrrrrr | rrrrr }
	\toprule
	& \textbf{Cora} & \textbf{CiteSeer} & \textbf{PubMed} & \textbf{Computers} & \textbf{Photo} & \textbf{ogbn-arXiv} & \textbf{ogbn-Products} & \textbf{Chameleon} & \textbf{Squirrel} & \textbf{Actor} & \textbf{Twitch} & \textbf{Pokec} \\
	\midrule
	    \# of Nodes & 2,708 & 3,327 & 19,717 & 13,752 & 7,650 & 169,343 & 2,449,029 & 2,277 & 5,201 & 7,600 & 168,114 & 1,632,803 \\
	    \# of Edges & 5,429 & 4,732 & 44,338 & 245,861 & 119,081 & 1,166,243 & 61,859,140 & 36,101 & 216,933 & 29,926 & 6,797,557 & 30,622,564 \\
	    \# of Features & 1433 & 3703 & 500 & 767 & 745 & 128 & 100 & 2325 & 2089 & 931 & 7 & 65 \\
	    \# of Classes & 7 & 6 & 3 & 10 & 8 & 40 & 39 & 5 & 5 & 5 & 2 & 2 \\
	\bottomrule
\end{tabular}}}
\label{table:datastat}
\end{table}

\subsection{Experimental Settings}
For fair comparison, each experiment is run with $5$ different splits of both synthetic and real-world datasets.
In link prediction, edges are split into training, validation and testing sets with the $70\%/10\%/20\%$ ratio.
In node classification, the nodes are split into training, validation and testing sets with the $2.5\%/2.5\%/95\%$ ratio.
For small graphs, the linear models are trained by L-BFGS for $100$ epochs with the patience of $5$, and the non-linear models are trained by ADAM for $1000$ epochs with the patience of $200$.
For large graphs (Products, Twitch, and Pokec), most models are trained by ADAM for $100$ epochs with the patience of $10$, except GPR-GNN and GAT, they are trained by ADAM for $20$ epochs with the patience of $5$ to speedup. 
All the training are full-batch, and the same amount of negative edges are randomly sampled for each batch while training.

\subsection{Hyperparameters} \label{app:hp}

\begin{wrapfigure}{L}{0.5\textwidth}
\vspace{-2mm}
\begin{minipage}[h] {1\linewidth}
\captionof{table}{Search space of hyperparameters. \label{tab:hyper}}
\centering{\resizebox{1\columnwidth}{!}{
    \begin{tabular}{l|l}
      \toprule
      \textbf{Method} & \textbf{Hyperparameters} \\
      \midrule
      GCN & $lr=[0.001, 0.01], wd=[0, 5e^{-4}], layers=2$ \\
      SAGE & $lr=[0.001, 0.01], wd=[0, 5e^{-4}], layers=2$ \\
      H$_2$GCN & $lr=[0.001, 0.01], wd=[0, 5e^{-4}], layers=[1, 2]$ \\
      GPR-GNN & $lr=[0.001, 0.01], wd=[0, 5e^{-4}], layers=10, \alpha=[0.1, 0.2, 0.5, 0.9]$ \\
      GAT & $lr=[0.001, 0.01], wd=[0, 5e^{-4}], layers=2, heads=[8, 16]$ \\
      \textsc{SlimG} & $lr=0.1, wd_{1}=[1e^{-4}, 1e^{-5}], wd_{2}=[1e^{-3}, 1e^{-4}, 1e^{-5}, 1e^{-6}]$ \\
      \midrule
      \method & $lr=0.1, wd_{1}=[1e^{-4}, 1e^{-5}], wd_{2}=[1e^{-3}, 1e^{-4}, 1e^{-5}, 1e^{-6}]$ \\
      \bottomrule
    \end{tabular}
}}
\end{minipage}
\vspace{-3mm}
\end{wrapfigure}

The search space of the hyperparameters is provided in Table~\ref{tab:hyper}.
Each experiment is run with $5$ different splits of the dataset, and grid search of hyperparameters based on the validation performance is done on each of the splits.
The hidden size of all methods is set to $128$.

\textbf{Linear GNNs.}
For sparse-group LASSO, $wd_{1}$ is the coefficient of overall sparsity, and $wd_{2}$ is the one of group sparsity.
The derived embedding $\mathbf{R}$ in \method uses $T=1000$ for the synthetic and OGB datasets, and $T=200$ for the real-world datasets.
The sample size $S$ is set to be $200,000$ in all experiments.
Since the large graphs (Products, Twitch, and Pokec) have no more than $128$ features, to ensure \method and \textsc{SlimG} have enough parameters, we concatenate the one-hot node degree to the original features.
The hidden size is set to $256$ for \method and \textsc{SlimG} in the OGB experiments.
For ogbl-ddi, since there is no node features, they use the one-hot node degree as the node features.
For ogbl-ppa, we concatenate the embedding from node2vec \citep{grover2016node2vec} to the original features, as \citet{hamilton2017inductive} did for GCN and SAGE.

% \clearpage
% \tableofcontents

\end{document}